%% file: 0_main.tex
\documentclass[nohyperref]{article}

\usepackage{microtype}
\usepackage{graphicx}
\usepackage{subfigure}
\usepackage{booktabs} 

\usepackage{hyperref}

\input{_packages}
\usepackage{stackengine}

\usepackage[accepted]{icml2022}

\usepackage{amsmath}
\usepackage{amssymb}
\usepackage{mathtools}
\usepackage{amsthm}

\theoremstyle{plain}

\theoremstyle{definition}

\theoremstyle{remark}

\icmltitlerunning{IGLUE: A Benchmark for Transfer Learning across Modalities, Tasks, and Languages}

\newcolumntype{Y}{>{\centering\arraybackslash}X}

\begin{document}

\twocolumn[
\icmltitle{IGLUE: A Benchmark for Transfer Learning \\across Modalities, Tasks, and Languages}



\icmlsetsymbol{equal}{*}

\begin{icmlauthorlist}
\icmlauthor{Emanuele Bugliarello}{cop,mila}
\icmlauthor{Fangyu Liu}{cam}
\icmlauthor{Jonas Pfeiffer}{tuda,nyu}
\icmlauthor{Siva Reddy}{mila,mcgill}
\icmlauthor{Desmond Elliott}{cop}
\icmlauthor{Edoardo Maria Ponti}{mila,mcgill}
\icmlauthor{Ivan Vulić}{cam}
\end{icmlauthorlist}

\icmlaffiliation{cop}{University of Copenhagen}
\icmlaffiliation{cam}{University of Cambridge}
\icmlaffiliation{tuda}{TU Darmstadt}
\icmlaffiliation{nyu}{New York University}
\icmlaffiliation{mila}{Mila -- Quebec Artificial Intelligence Institute }
\icmlaffiliation{mcgill}{McGill University}

\icmlcorrespondingauthor{Emanuele Bugliarello}{emanuele@di.ku.dk}

\icmlkeywords{Multimodal Learning, vision and language, multilingual, benchmark}

\vskip 0.3in
]



\printAffiliationsAndNotice{}  

\begin{abstract}
Reliable evaluation benchmarks designed for replicability and comprehensiveness have driven progress in machine learning. Due to the lack of a multilingual benchmark, however, vision-and-language research has mostly focused on English language tasks. To fill this gap, we introduce the Image-Grounded Language Understanding Evaluation benchmark. IGLUE brings together---by both aggregating pre-existing datasets and creating new ones---visual question answering, cross-modal retrieval, grounded reasoning, and grounded entailment tasks across 20 diverse languages. Our benchmark enables the evaluation of multilingual multimodal models for transfer learning, not only in a zero-shot setting, but also in newly defined few-shot learning setups. Based on the evaluation of the available state-of-the-art models, we find that translate-test transfer is superior to zero-shot transfer and that few-shot learning is hard to harness for many tasks. Moreover, downstream performance is partially explained by the amount of available unlabelled textual data for pretraining, and only weakly by the typological distance of target--source languages. We hope to encourage future research efforts in this area by releasing the benchmark to the community.
\end{abstract}

\section{Introduction}
\label{s:introduction}
\input{01_intro}

\section{Related Work and Motivation}
\label{s:rw}
\input{02_rw}

\section{IGLUE Benchmark}
\label{s:benchmark}
\input{03_benchmark}

\section{Evaluation Framework}
\label{s:framework}
\input{04_framework}

\section{Main Results and Discussion}
\label{s:results}
\input{05_results}

\section{Conclusion and Outlook}
\label{s:conclusion}
\input{07_conclusion}

\section*{Acknowledgements}
{\scriptsize\euflag} We are grateful to Heather Lent, Semih Yagcioglu, Ilker Kesen, and Mustafa Sercan Amac for their constructive feedback and help. We also thank Krishna Srinivasan for providing us with an early release of the WIT test data and further clarifications. This project has received funding from the European Union's Horizon 2020 research and innovation programme under the Marie Sk\l{}odowska-Curie grant agreement No 801199 (\emph{Emanuele Bugliarello}).
The work of \emph{Fangyu Liu} has been supported by Grace \& Thomas C.H. Chan Cambridge Scholarship.
The work of \emph{Jonas Pfeiffer} has been supported by the LOEWE initiative (Hesse, Germany) within the emergenCITY center.
The work of \emph{Ivan Vuli\'{c}} has been supported by the ERC PoC Grant MultiConvAI (no. 957356), and a research donation from Huawei.
The work of \emph{Edoardo Ponti} and \emph{Siva Reddy} has been supported by the Facebook CIFAR AI Chair program.
The annotation cost is partially funded by Cambridge Digital Humanities Digitisation/Digital Resources Awards and the NSERC Discovery Grant.

\bibliography{custom}
\bibliographystyle{icml2022}

\clearpage
\appendix

\input{08_app}

\end{document}

%% file: _packages.tex
\usepackage{microtype}
\usepackage{booktabs} 
\usepackage{mathtools}
\usepackage{tikz}
\usetikzlibrary{bayesnet}
\usetikzlibrary{arrows}
\usepackage[noabbrev,capitalise]{cleveref}
\usepackage{tabularx}
\usepackage{tabulary}
\usepackage{colortbl}
\usepackage{xspace}
\usepackage{float}
\newcommand\doublecheck{{\checked\kern-0.5em\checked}}

\usepackage{fontawesome5}
\newcommand\checked{\footnotesize \faIcon{check}}

\usepackage{mathtools}
\newcounter{tableeqn}[table]

\newcounter{tablesubeqn}[tableeqn]

\usepackage{relsize}
\usepackage{amsmath, amsthm, amssymb}
\usepackage{mathrsfs}

\usepackage{caption}
\usepackage{graphicx}

\usepackage{enumitem}

\usepackage{color, colortbl}
\PassOptionsToPackage{table,dvipsnames}{xcolor}
\usepackage{multirow,makecell}
\usepackage{rotating}
\usepackage{booktabs}
\usepackage[table]{xcolor}
\usepackage{tabularx}

\usepackage{todonotes}
\makeatletter
\newcommand*\iftodonotes{\if@todonotes@disabled\expandafter\@secondoftwo\else\expandafter\@firstoftwo\fi}
\makeatother

\definecolor{edolime}{RGB}{250,218,94}

\usepackage{euflag}

\newcommand{\band}{\rowcolor{gray!10}}

\newcommand{\rparagraph}[1]{\vspace{0.0mm}\noindent\textbf{#1.}}
\newcommand{\benchmark}{{\sc IGLUE}\xspace}
\newcommand{\vl}{V\&L\xspace}
\newcommand{\volta}{{\sc Volta}\xspace}

\newcommand{\english}{\textsc{eng}}
\newcommand{\arab}{\textsc{arb}}
\newcommand{\bengali}{\textsc{ben}}
\newcommand{\bulgarian}{\textsc{bul}}
\newcommand{\chinese}{\textsc{cmn}}
\newcommand{\danish}{\textsc{dan}}
\newcommand{\estonian}{\textsc{est}}
\newcommand{\french}{\textsc{fra}}
\newcommand{\german}{\textsc{deu}}
\newcommand{\greek}{\textsc{ell}}
\newcommand{\indonesian}{\textsc{ind}}
\newcommand{\japanese}{\textsc{jpn}}
\newcommand{\korean}{\textsc{kor}}
\newcommand{\portuguese}{\textsc{por}}
\newcommand{\russian}{\textsc{rus}}
\newcommand{\spanish}{\textsc{spa}}
\newcommand{\swahili}{\textsc{swa}}
\newcommand{\tamil}{\textsc{tam}}
\newcommand{\turkish}{\textsc{tur}}
\newcommand{\vietnamese}{\textsc{vie}}

\newcommand{\czech}{\textsc{ces}}

%% file: 01_intro.tex
Until recently, advances in multimodal vision-and-language (\vl) modelling have predominantly focused on English or a few high-resource Indo-European languages~\citep{elliott-etal-2016-multi30k}, disregarding the enormous diversity of the world's languages and cultures \cite{Bender_2011,ponti-etal-2019-modeling,liu-etal-2021-visually}, which exacerbated the Anglo-centric bias \cite{Jauhar:2018arxiv,ponti-etal-2020-xcopa} of \vl models.

Nonetheless, this trend is reversing by virtue of a series of independent efforts in the the \vl research community: on top of expanding multilingual image--sentence retrieval to lower-resource languages \cite{10.1145/3404835.3463257} and additional modalities~\cite{10.1145/3340531.3412783}, we have recently witnessed pioneering work on multimodal machine translation \citep{yao-wan-2020-multimodal,huang-etal-2020-unsupervised-multimodal}, multilingual visual question answering \citep{pfeiffer2021xgqa}, multilingual text-to-video search \citep{huang-etal-2021-multilingual,lei-etal-2021-mtvr}, multilingual visual reasoning \citep{liu-etal-2021-visually}, and cross-modal retrieval \citep{jain-etal-2021-mural-multimodal}. 
Yet, multilingual multimodal models~\citep{Ni_2021_CVPR,jain-etal-2021-mural-multimodal} are typically tested only on the task of cross-modal retrieval.

\begin{figure*}[t]
 	\centering
 	\includegraphics[width=0.85\linewidth, trim={0cm 6.8cm 3cm 0cm}, clip]{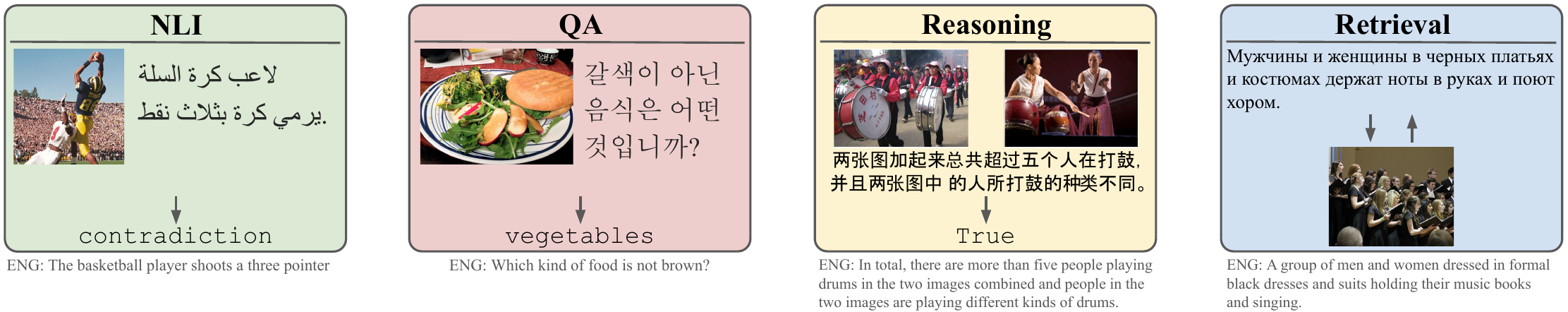}%
 	\vspace{-0.1cm}
 	\caption{Overview of the tasks in \benchmark, which include grounded natural language inference, visual question answering, grounded reasoning, and cross-modal retrieval. Each task is associated with an example of input and output (English translations at the bottom).}
 	\label{fig:tasks}
 	\vspace{-0.3cm}
\end{figure*}

In order to incentivise and guide future research in multilingual \vl research, we introduce the \textbf{I}mage-\textbf{G}rounded \textbf{L}anguage \textbf{U}nderstanding \textbf{E}valuation (\benchmark) benchmark. 
We create \benchmark by collating current research threads in this area and extending them with two datasets for cross-lingual visual entailment (XVNLI) and image--text retrieval (xFlickr\&CO), for an even more comprehensive coverage of tasks.
\benchmark is the first evaluation suite for multilingual multitask \vl modelling, comprising five datasets across four structurally different tasks that require different levels of syntactic-semantic \vl understanding in cross-lingual setups: cross-modal retrieval, visual question answering, natural language inference, and visual reasoning (\cref{fig:tasks}). 
Aiming at a representative selection of languages, \benchmark covers 20 typologically diverse languages, spanning 11 language families, 9 scripts, and 3 WALS-defined \cite{wals} geographical macro-areas.

Similar to previous text-based, cross-lingual transfer benchmarks, such as XGLUE \citep{liang-etal-2020-xglue} and XTREME \citep{pmlr-v119-hu20b}, our \vl benchmark is designed to evaluate zero-shot transfer scenarios, where annotated training data is provided in English, but none in the target language. 
In addition, contrary to the above-mentioned datasets, \benchmark also provides standardised data splits to guide cross-lingual few-shot learning experiments, which can help reduce the gap between zero-shot and supervised performance \cite{lauscher-etal-2020-zero,zhao-etal-2021-closer}. 
In this setup, a small number of task-annotated examples for fine-tuning are available for a given target language. 
We also release machine-translated versions of the test sets to enable the evaluation of `translate test' cross-lingual transfer.

By virtue of the newly created IGLUE benchmark, we also run the first systematic comparative evaluation of
cutting-edge multilingually pretrained \vl architectures \cite{Ni_2021_CVPR,Zhou_2021_CVPR,liu-etal-2021-visually}, as well as a series of representative monolingual \vl models combined with machine translation \citep[\textit{inter alia}]{NEURIPS2019_c74d97b0,10.1007/978-3-030-58577-8_7} across a range of diverse \vl tasks and languages. 

This evaluation offers new insights on the models' core strengths and current limitations. 
Foreshadowing, we showcase large gaps between performance in English and transfer performance, where (to a limited extent) the gaps are more prominent in lower-resource languages and languages more distant from English.
In addition, leveraging target-language in-task few shots is remarkably arduous: whereas in previous text-only experiments \cite{lauscher-etal-2020-zero}, \textit{ceteris paribus} there are huge gains of few-shot scenarios over their zero-shot counterparts, we demonstrate that current \vl models often require thousands of examples before showing signs of improvement. 
Finally, performance also seems correlated with task difficulty: for instance, NLI shows the smallest gaps between English and other languages and QA benefits the most from few-shot adaptation. 
On the other hand, visually grounded reasoning and cross-modal retrieval appear to be harder in both respects.

\rparagraph{Contributions} 
\textbf{1)} In order to guide and inspire more work in the area of multilingual \vl research, we present a first evaluation benchmark for cross-lingual transfer learning for \vl tasks, spanning 20 languages, 5 datasets, and 4 different tasks. \textbf{2) } In the process of benchmark creation, we complement existing datasets with new training and evaluation data in several languages to increase diversity and enable few-shot learning, and introduce a first multilingual dataset for visually grounded cross-lingual NLI. \textbf{3)} We conduct systematic evaluations of representative \vl architectures in zero-shot and few-shot cross-lingual transfer scenarios, offering standard data splits and empirical baselines for future research. \textbf{4)} Our results and additional analyses take stock of the current gaps and challenges in cross-lingual \vl research. \textbf{5)} To further facilitate replicable research in this area, we re-implement the existing multilingual \vl pretrained encoders in a unified framework (\volta; \citealt{bugliarello-etal-2021-multimodal}), which also provides access to five English \vl BERTs and 12 \vl tasks. 
We provide data and code for the evaluation of multilingual \vl models at \url{https://iglue-benchmark.github.io/}.

%% file: 02_rw.tex
\rparagraph{Multilingual Multimodal Learning}
Multilingual \vl research focuses on collecting resources, developing models, and evaluating systems that need to jointly reason over multilingual text and multimodal inputs, this way combining two areas of crucial importance: multilingual \cite{10.5555/3104322.3104328,ponti-etal-2019-modeling} and multimodal learning \cite{10.5555/3013558.3013571,10.1109/TPAMI.2018.2798607}. 
A natural overlap of the two areas is \textit{language grounding} in perception \citep[typically vision;][]{5206848,kiela-etal-2018-learning}, where the perceptual input can be considered as an inherent cross-lingual signal \cite{kiela-etal-2015-visual,gella-etal-2017-image,caglayan-etal-2021-cross}. 
Many other research goals lie at the intersection of these two areas, such as transfer learning and modularity \citep{ponti-etal-2021-parameter,ansell2021composable}. 

However, until recently the pace of progress in multilingual multimodal learning has not gained as much momentum as \vl in monolingual settings, mostly due to the scarcity of resources for training and evaluation. 
Texts in most multimodal datasets are usually only available in English, or in high-resource languages \citep[e.g., Chinese and a few major Indo-European languages;][]{elliott-etal-2016-multi30k,elliott-etal-2017-findings,barrault-etal-2018-findings,Wang_2019_ICCV}. 
Consequently, models trained on such datasets do not take into account linguistic diversity \cite{ponti-etal-2020-xcopa} or cross-cultural nuances \cite{10.1145/3340531.3412783,liu-etal-2021-visually,yin-etal-2021-broaden}.\footnote{Some notable exceptions, which did reach beyond English, were still limited to the image--text retrieval task and mostly Indo-European languages \cite{rotman-etal-2018-bridging,9010736,huang-etal-2019-multi,Kim_Saito_Saenko_Sclaroff_Plummer_2020,su-etal-2021-gem}.}

The need to expand \vl research towards more languages has been recognised by \textbf{1)} the recent creation of multilingual training and evaluation data across diverse \vl tasks and languages \citep[\textit{inter alia}]{10.1145/3340531.3412783,10.1145/3404835.3463257,su-etal-2021-gem,pfeiffer2021xgqa,liu-etal-2021-visually,Wang2021MultiSubs}, as well as \textbf{2)} the emergence of the first large multilingual-multimodal pretrained models \cite{Ni_2021_CVPR,Zhou_2021_CVPR,liu-etal-2021-visually} and monolingual \vl models adapted to multiple languages \cite{10.1007/978-3-030-58577-8_7,pfeiffer2021xgqa}. 
In this work, we merge and expand on these two research threads, aiming to highlight current achievements and challenges in this area and to facilitate comparative evaluations, thus bringing together the above-mentioned collective research efforts.

\rparagraph{Multi-Task Evaluation Benchmarks in NLP}
\benchmark has been inspired by recent text-only \textit{multi-task} benchmarks for natural language understanding: these benchmarks have been proven invaluable as key drivers of recent steep performance progress of NLP. The creation of such benchmarks in monolingual settings, sparked by the pioneering and now omnipresent English-only GLUE \citep{wang-etal-2018-glue} and SuperGLUE \citep{NEURIPS2019_4496bf24}, has been extended to other languages such as Indonesian \cite{wilie-etal-2020-indonlu}, Korean \cite{park2021klue}, Russian \cite{shavrina-etal-2020-russiansuperglue}, and Romanian \cite{dumitrescu2021liro}. 

\benchmark is even more related to multi-task benchmarks developed for \emph{cross-lingual transfer settings}: XTREME \cite{pmlr-v119-hu20b}, XGLUE \citep{liang-etal-2020-xglue}, and XTREME-R \cite{ruder-etal-2021-xtreme}. They have brought in the spotlight the necessity to evaluate not only on a diverse set of tasks, but also on a diverse set of languages, in order to incentivise research on models that forgo monolingual limitations and generalise well in multilingual settings.

With \benchmark, we make the first leap into multimodal cross-lingual evaluation. Driven by analyses of text-only benchmarks and lessons learned from them \cite{ethayarajh-jurafsky-2020-utility}, \benchmark aims to extend sheer accuracy-driven evaluation towards other crucial aspects such as fine-tuning efficiency, sample efficiency and adaptation to low-data scenarios, and enhance multilingual inclusivity and diversity.

%% file: 03_benchmark.tex
\textbf{Design Principles.}
Aiming for a comprehensive resource, we take inspiration from the best practices and design principles of existing multilingual benchmarks \citep{pmlr-v119-hu20b,liang-etal-2020-xglue,ruder-etal-2021-xtreme}, and adapt them to account for the unique challenges of \vl tasks. In particular, we abode by the following principles: 

\noindent \emph{(1) Task Diversity}. We selected a wide spectrum of multimodal tasks that reflect multiple facets of \vl learning.

\noindent \emph{(2) Language Diversity}. The datasets included in \benchmark should cover many languages that are diverse in terms of family, geographic area, typological features, and script. 

\noindent \emph{(3) Accessibility}. The data should be released under a licence that permits use and redistribution for research purposes. 

Compared to previous text-only benchmarks, \vl tasks are more time- and compute-intensive by nature.
To widen usability, we created \benchmark by carefully choosing datasets and training setups that would enable quick development even by practitioners constrained by limited resources. 

\newcolumntype{R}{>{\raggedleft\arraybackslash}X}
\newcolumntype{Y}{>{\centering\arraybackslash}X}
\newcolumntype{Z}{>{\raggedright\arraybackslash}X}
\begin{table}[t]
\caption{\benchmark statistics. $^\dagger$Aggregated across languages. Language counts include English. MaxS: Max-Shot. F: Flickr; VG: Visual Genome; N: NLVR2; M: MaRVL; C:COCO; W: Wikipedia}
\label{tab:taskstats}
\setlength\tabcolsep{2pt}
\center
\def\arraystretch{0.9}
\small
  \begin{tabularx}{\columnwidth}{l RRRrr}
\toprule
\textbf{Task} &  \multicolumn{1}{c}{\textbf{NLI}}   & \multicolumn{1}{c}{\textbf{QA}} & \multicolumn{1}{c}{\textbf{Reason}}  & \multicolumn{2}{c}{\textbf{Retrieval}} \\
  \cmidrule(r){2-2}
  \cmidrule(r){3-3}
  \cmidrule(r){4-4}
  \cmidrule(r){5-6}
Train/Dev data & SNLI & GQA & NLVR2 & \multicolumn{1}{r}{Flickr30K} & \multicolumn{1}{r}{WIT$_{\textsc{en}}^{500\text{K}}$} \\
Test data & \multicolumn{1}{Y}{XVNLI} & \multicolumn{1}{Y}{xGQA} & \multicolumn{1}{Y}{MaRVL} & \multicolumn{1}{r}{xFlickr\&CO} & \multicolumn{1}{r}{WIT} \\
Languages & 5 & 8 & 6 & 8 & 11 \\
Img source & F & VG & N + M & F + C & W \\
Train images & 30K & 72K & 87K & 29K & 469K \\
Train samples & 541K & 943K & 86K & 145K & 500K \\
Dev images & 1K & 10K & 7K & 1K & 4.6K \\
Dev samples & 18K & 132K & 8K & 5K & 4.6K \\
Test images & 357 & 300 & $^\dagger$4.9K & 2K & $^\dagger$6.2K \\
Test samples & 1.1K & 9.6K & $^\dagger$5.7K & 2K & $^\dagger$9.6K \\
MaxS images & 48 & 48 & 80 & 100 & - \\
MaxS samples & 168 & 1.5K & 80 & 100 & - \\
\bottomrule
\end{tabularx}
\vspace{-0.5cm}
\end{table}

\subsection{Tasks and Datasets}
Challenges in \vl learning are multi-faceted and entangled with distinct abilities: e.g., drawing inferences, reasoning over and comparing images, answering questions, and retrieving images or their captions. In \benchmark we thus represent all these abilities together with their corresponding tasks, see~\cref{tab:taskstats} and~\cref{fig:tasks} for an overview.
We opt for the following \vl tasks based on (i) (partial) availability of multilingual data, and (ii) computational requirements.

\rparagraph{XVNLI}
We propose the new task of Cross-lingual Visual Natural Language Inference (XVNLI). It requires the model to predict if a text-\textit{hypothesis} `entails', `contradicts', or is `neutral' to an image-\textit{premise}.  
We combine the text-only dataset  SNLI \cite{bowman-etal-2015-large}, with its multimodal \cite{Xie2019NLIVE} and cross-lingual \cite{agic-schluter-2018-baselines} counterparts.\footnote{\citet{vu-etal-2018-grounded} formulate grounded textual entailment as the task where the image supplements the textual premise and hypothesis. We follow the established setup of \citet{Xie2019NLIVE}.}
We provide new train, development, and test splits such that the test split consists of images covered by the underlying cross-lingual text-only dataset. 
We discard all text examples of images in the test set for which no cross-lingual version exists.
The remaining images for which only English hypothesis-examples exist are randomly split into the train and development set. 
To mitigate data leakage between the splits, we sample based on the images; all text examples corresponding to one image are added to one split. 

\rparagraph{xGQA} 
To solve the Cross-lingual Grounded Question Answering task (xGQA; \citealt{pfeiffer2021xgqa}), a model must answer several types of \emph{structured} questions about an image. 
The evaluation data in 7 target languages are manually translated from the validation set of GQA \citep{Hudson_2019_CVPR}, whereas the training data are sourced from the English training set of GQA. 
All images were originally sampled from Visual Genome \citep{10.1007/s11263-016-0981-7}.
In particular, we use the English \emph{balanced} training set to train our models, and evaluate on the \emph{few-shot} evaluation sets defined by \citet{pfeiffer2021xgqa} to allow for direct comparison between zero-shot and few-shot experiments.

\rparagraph{MaRVL}
The Multicultural Reasoning over Vision and Language dataset \citep{liu-etal-2021-visually} requires to determine whether a textual description is \textit{true} or \textit{false} about a pair of images. This involves comparing their visual representations and reasoning about the facts in the description. The creation of MaRVL is entirely driven by native speakers. As a consequence, the descriptions are written from scratch and images are selected to be culturally relevant. The NLVR2 data \citep{suhr-etal-2019-corpus} in English are used for training.

\rparagraph{xFlickr\&CO} 
We create a new, multilingual evaluation set for retrieval by combining 1{,}000 images from the Flickr30K~\citep{young-etal-2014-image} and 1{,}000 from the COCO~\citep{10.1007/978-3-319-10602-1_48} test split defined by~\citet{7298932} (henceforth, \emph{Karpathy split}). Each image is then associated with a single caption.\footnote{We expand the Flickr30K evaluation set since state-of-the-art model performance is currently close to the ceiling of human performance \citep{pmlr-v139-jia21b}. We show that this leads to a more challenging evaluation set which requires out-of-distribution generalisation, while maintaining an efficient training regime.} 
The English evaluation set is obtained by sampling one caption per image from the existing datasets. 
With the exception of Japanese~\citep{yoshikawa-etal-2017-stair,nakayama-etal-2020-visually}, no other language covers the evaluation splits of both Flickr and COCO.\footnote{Multilingual captions for Flickr30K also exist in Dutch~\citep{van-miltenburg-etal-2018-didec} and Mandarin~\citep{10.1145/3123266.3123366}, but the Mandarin captions do not match the expected test split. COCO has also been (partially) translated to Hindi~\citep{10.1145/3432246}, Italian~\citep{IJCOL:scaiella_et_al:2019}, Mandarin~\citep{8630050}, Spanish~\citep{garcia-2020-mscocoes} and Vietnamese~\citep{10.1007/978-3-030-63007-2_57}, with similar problems. The Mandarin captions have even been used to train and evaluate M$^3$P and UC$^2$ in a multilingual setup. We find that 882/1{,}000 test images are used as training examples in other languages, and that 148 of its training images are in the expected test split. In a similar vein, we find that 946/1{,}000 test samples in XTD~\citep{aggarwal2021towards}---which contains COCO translations in seven languages (Italian, Korean, Mandarin, Portuguese, Russian, Spanish and Turkish)---are from the COCO training split, while only 7 of them are also in the Karpathy COCO test split. These overlaps lead to considerable leakage of visual and cross-lingual information during fine-tuning and evaluation.} 
We hence crowdsource image descriptions in 6 other languages by asking annotators to describe each image, rather than translating its English caption, to prevent biasing their description~\citep{van-miltenburg-etal-2017-cross,frank_elliott_specia_2018}.
We ask annotators to follow the Flickr30K guidelines (see \cref{sec:xflickrco} for details).
For the Flickr30K subsplit in German, we sample captions from the Multi30K task 2 data \citep{elliott-etal-2016-multi30k}.

\rparagraph{WIT} The Wikipedia-based Image Text dataset \citep{10.1145/3404835.3463257} collected examples from Wikipedia in 108 languages. 
Similar to xFlickr\&CO, the tasks consist in retrieving the correct image given a textual description (image retrieval) and vice versa (text retrieval). 
WIT represents a very diverse set of concepts and real world entities. 
Text fields in WIT often tend to be descriptive, verbose and use specific terminology, different from single-line captions of common words and objects in Flickr30K and COCO.
This allows us to evaluate the high-level image--sentence understanding of multilingual \vl models across a wider range of languages and real-world entities than in xFlickr\&CO.
For training, we randomly sample a subset of 500K captions from the English training set of WIT.
For evaluation, we use the WIT test data released as part of its corresponding Kaggle competition.\footnote{\url{www.kaggle.com/c/wikipedia-image-caption}.}
In particular, as most of the test languages are Indo-European, we choose the 4 national languages with the lowest Wikipedia coverage (measured in number of articles as of Jan 2022), and 6 more languages to both cover different properties (language families, scripts) and overlap with other tasks in \benchmark.\footnote{While~\citet{jain-etal-2021-mural-multimodal} define a set of 8 low-resource languages for evaluation on WIT, they were sampled from the training set. Moreover, we find that several of their test images contain pictures with identifiable people, which would complicate its public release due to regulations such as the GDPR in Europe. By only relying on the true test split, we allow researchers to easily adopt the full WIT training dataset in their experimental pipelines.}
We ensure that each language has at least 500 image--caption pairs, and limit English and Japanese data to 1{,}000 samples (see also~\cref{sec:WIT}). 

\subsection{Languages}

\begin{table}[t]
\caption{Benchmark languages and tasks. English is only used used for training. Tasks legend: \doublecheck{} train and test sets available; \checked{} test-only data available; * Japanese captions in xFlickr\&CO are translations from English. Languages legend: A: Asiatic; C: Congo; E: European; T: Tibetan; Austron: Austronesian. Language codes are based on the ISO 639-3 international standard.} 
\label{tab:langs-tasks}
\vskip 0.15in
\def\arraystretch{0.95}
\setlength\tabcolsep{1pt}
\small
\center
  \resizebox{0.48\textwidth}{!}{
  \begin{tabular}{llllccccc}
\toprule
\multicolumn{4}{c}{\textbf{Language}} & \textbf{NLI} & \textbf{QA} & \textbf{Reasoning}  & \multicolumn{2}{c}{\textbf{Retrieval}} \\
  \cmidrule(r){1-4}
  \cmidrule(r){5-5}
  \cmidrule(r){6-6}
  \cmidrule(r){7-7}
  \cmidrule(r){8-9}
Name & Code & Family & Script & XVNLI & xGQA & MaRVL & xFlickr\&CO & WIT \\
\midrule
\band English & \english & Indo-E & Latin & \doublecheck{} & \doublecheck{} & \doublecheck{} & \doublecheck{} & \doublecheck{} \\
\hline
Arabic & \arab & Afro-A & Arabic & \doublecheck{} & & & & \checked{} \\
Bengali & \bengali & Indo-E  & Bengali & & \doublecheck{}        &      &             &            \\
Bulgarian & \bulgarian & Indo-E  & Cyrillic & &         &      &              & \checked{}           \\
Danish & \danish & Indo-E  & Latin & &         &      &              & \checked{}           \\
Estonian & \estonian & Uralic  & Latin & &         &      &              & \checked{}           \\
German & \german & Indo-E & Latin & & \doublecheck{}        &      & \doublecheck{}             &            \\
Greek & \greek & Indo-E & Greek & &         &      &             & \checked{}           \\
French & \french & Indo-E & Latin & \doublecheck{}      &         &      &             &            \\
Indonesian & \indonesian & Austron & Latin & & \doublecheck{}        & \doublecheck{}     & \doublecheck{} & \checked{} \\
Japanese & \japanese & Japonic & Kanji & &         &      & *\doublecheck{}~~             & \checked{}           \\
Korean & \korean & Koreanic & Hangul & & \doublecheck{}        &      &             & \checked{}           \\
Mandarin & \chinese & Sino-T & Hanzi & & \doublecheck{}        & \doublecheck{}     & \doublecheck{} & \\
Portuguese & \portuguese & Indo-E & Latin & & \doublecheck{}        &      &             &            \\
Russian & \russian & Indo-E & Cyrillic & \doublecheck{}      & \doublecheck{}        &      & \doublecheck{}             & \\
Spanish & \spanish & Indo-E & Latin & \doublecheck{}      &         &      & \doublecheck{} & \\
Swahili & \swahili & Niger-C & Latin &       &         & \checked{}     &             &            \\
Tamil & \tamil & Dravidian & Tamil & &         & \checked{}     &             &            \\
Turkish & \turkish & Turkic & Latin & &         & \doublecheck{}     & \doublecheck{}             & \checked{}           \\
Vietnamese & \vietnamese & Austro-A & Latin & &         &      &             & \checked{}           \\
\bottomrule
\end{tabular}
}
\vspace{-0.2cm}
\end{table}

\benchmark covers 20 languages (\cref{tab:langs-tasks}).
They are extremely diverse, as they comprise 11 distinct language families, and span across 3 of the 5 geographic macro-areas defined by \citet{wals}.
Moreover, they are written in a variety of scripts: Arabic, Bengali--Assamese, Chinese characters, Cyrillic, Greek, Hangul, Kanji, Latin, and Tamil.
The distribution of languages per task is also shown in~\cref{tab:langs-tasks}.

We ensured that \benchmark includes languages with different levels of unlabelled data available \citep{joshi-etal-2020-state,blasi2021systematic}. 
Thus, it allows for evaluating models with different data paucity regimes during pretraining. 
While not all languages are covered in each task---due to the (partly) independent selection of languages in the original datasets---it is worth noting that 10 out of 20 languages have data for two or more tasks. 
Thus, \benchmark might also facilitate future research in cross-task knowledge transfer.

\subsection{From Zero-Shot to Few-Shot Setups}
The established practice in multilingual text-based benchmarks \citep{pmlr-v119-hu20b,liang-etal-2020-xglue,ruder-etal-2021-xtreme} is to frame cross-lingual transfer as a \textit{zero-shot} learning problem. 
However, multilingual pretrained models can be additionally fine-tuned in a \textit{few-shot} learning setup; that is, on a handful of data points in a target language annotated for a specific task.
While computationally inexpensive, this strategy has been proven as very beneficial for performance in text-only tasks, especially on low-resource and distant languages \citep{lauscher-etal-2020-zero,ponti-etal-2021-minimax}.

We hence extend IGLUE to support few-shot learning setups by collecting samples for each \vl task.
In doing so, we use the notion of `annotation context' to define what a `shot' means in each dataset: For instance, this corresponds to an image and its caption in cross-modal retrieval, and to an image and all its questions in visual question answering.
For more details and statistics, we refer to~\cref{sec:datasets}.

Notably, the set of examples used for few-shot cross-lingual transfer may vary across experiments. 
The lack of a standard set for few-shot learning harms the replicability and comparability of such experiments, given that model performance exhibits a strong sensitivity to the selection of few-shot examples \citep{zhao-etal-2021-closer}. 
Therefore, following \citet{schick-schutze-2021-just} and \citet{pfeiffer2021xgqa}, in addition to the train / validation / test splits, we also release standard few-shot splits for every task and language (\cref{sec:datasets}).

%% file: 04_framework.tex
The new IGLUE benchmark allows for running a series of unprecedented comparative experiments and analyses. Here, we provide an overview of our experimental setup.
We evaluate all the existing multilingual \vl pretrained models released so far.
In particular, in order to further facilitate research and development in multilingual \vl modelling, we re-implement them in a single framework based on \volta~\citep{bugliarello-etal-2021-multimodal}\footnote{\url{https://github.com/e-bug/volta}.} in PyTorch~\citep{NEURIPS2019_9015}, which also provides access to five English pretrained models and twelve downstream tasks.\footnote{Code to reproduce our results is available online at \url{https://github.com/e-bug/iglue}.}
For further details about the framework, we refer to \cref{sec:exp-details}.

\rparagraph{Pretrained Models}
The state-of-the-art multilingual \vl models that we evaluate follow a BERT-like architecture~\citep{devlin-etal-2019-bert}. 
Their input is the concatenation of image region features extracted with a Faster R-CNN~\citep{NIPS2015_5638} object detector and textual tokens~\citep{sennrich-etal-2016-neural,wu2016google}.
The inputs are then processed by a single stack of Transformer layers~\citep{NIPS2017_7181} to obtain multimodal, contextualised representations.
Each model is trained to minimise multiple objectives---with minimal variations across models---to learn how to understand text (masked language modelling), vision (masked region modelling) and their co-occurrence (image--text matching). We provide only condensed summaries here, and refer the reader to the original work for further details.
\begin{itemize}[noitemsep, topsep=1pt]
    \item \textbf{mUNITER and xUNITER.} \citet{liu-etal-2021-visually} extend the UNITER architecture~\citep{10.1007/978-3-030-58577-8_7} multilingually by following the base approach of \citet{Ni_2021_CVPR}: a batch of multimodal English data from Conceptual Captions (CC; \citealt{sharma-etal-2018-conceptual}) is alternated with a batch of text-only multilingual Wikipedia data (sampled from the 104 languages used for mBERT). 
    The two models mainly differ in their initialisation: mUNITER from mBERT~\citep{devlin-etal-2019-bert} and xUNITER from XLM-R~\citep{conneau-etal-2020-unsupervised}.
    \item \textbf{M$^3$P.} \citet{Ni_2021_CVPR} further introduce multimodal code-switched training tasks where words in English captions are randomly replaced with a translation with a certain probability. This setup allows the model to explicitly align images with non-English languages. In particular, the authors obtain word translations from bilingual dictionaries in 50 languages. In each multimodal batch, data from CC is fed to the model either fully in English or with code-switched words according to a given sampling ratio. The text-only multilingual batch is sampled from Wikipedia in 100 languages. The model is initialised from XLM-R.
    \item \textbf{UC$^2$.} \citet{Zhou_2021_CVPR} rely on (text-only) machine translation technologies to obtain CC data in five languages (Czech, French, German, Japanese, and Mandarin). The model is then solely pretrained on multilingual multimodal batches, where, for each image, a caption is sampled uniformly from the available languages. Two pretraining objectives are added to those above, to (i) tighten region-token matching and (ii) align translations in a similar vein as~\citet{NEURIPS2019_c04c19c2} and \citet{caglayan-etal-2021-cross}. The approach also maps the object detector space onto the language model space before multimodal pretraining. As xUNITER and M$^3$P, UC$^2$ is initialised from XLM-R.
\end{itemize}

In addition to evaluating all the released multilingual \vl encoders, we also benchmark representative English \vl encoders pretrained in a controlled setup~\citep{bugliarello-etal-2021-multimodal} in combination with `translate test' transfer:\footnote{We obtain translations to English via Google Translate.} LXMERT~\citep{tan-bansal-2019-lxmert}, UNITER~\citep{10.1007/978-3-030-58577-8_7}, ViLBERT~\citep{NEURIPS2019_c74d97b0}, VisualBERT~\citep{li2019visualbert} and VL-BERT~\citep{Su2020VL-BERT:}.

\begin{table}[t]
\caption{Performance of multimodal models trained and evaluated on English test splits of the \benchmark tasks.} \label{tab:english}
\vskip 0.05in
\def\arraystretch{0.9}
\setlength\tabcolsep{2pt}
\small
\center
  \resizebox{0.48\textwidth}{!}{
  \begin{tabular}{lrrcrrrr}
\toprule
\multirow{2}{*}{\small \shortstack{\textbf{Model}}}         &  \multicolumn{1}{c}{\textbf{NLI}}   & \multicolumn{1}{c}{\textbf{QA}} & \multicolumn{1}{c}{\textbf{Reasoning}}  & \multicolumn{4}{c}{\textbf{Retrieval}} \\
  \cmidrule(r){2-2}
  \cmidrule(r){3-3}
  \cmidrule(r){4-4}
  \cmidrule(r){5-8}
 & \multicolumn{1}{c}{XVNLI} & \multicolumn{1}{c}{xGQA}     & \multicolumn{1}{c}{MaRVL} & \multicolumn{2}{c}{xFlickr\&CO}          & \multicolumn{2}{c}{WIT}        \\
 & & & (NLVR2 Test-P) & \multicolumn{1}{c}{IR} & \multicolumn{1}{c}{TR} & \multicolumn{1}{c}{IR} & \multicolumn{1}{c}{TR} \\
\midrule
\band mUNITER & 76.38 & 54.68 & \textbf{71.91} & \textbf{44.50} & \textbf{40.90} & {19.90} & \textbf{22.34} \\
\band xUNITER & 75.77 & 54.83 & 71.55 & 38.45 & 32.05 & 16.70 & 18.54 \\
\band UC$^2$ & 76.38 & {55.19} & {70.56} & 37.40 & 34.55 & 17.90 & 19.71 \\
\band M$^3$P & {76.89} & 53.75 & 68.22 & 31.35 & 24.60 & 15.50 & 15.33 \\
\midrule
\band LXMERT & 76.72 & 53.04 & 68.95 & 36.85 & 29.60 & 18.00 & 18.50 \\
\band UNITER & 77.15 & 55.59 & 71.11 & 43.05 & 40.70 & 18.10 & 19.00 \\
\band ViLBERT & 77.15 & \textbf{55.82} & 71.02 & 38.70 & 35.70 & 17.90 & 20.20 \\
\band VisualBERT & 77.15 & 53.67 & 71.18 & 43.95 & 39.85 & 18.30 & 20.50 \\
\band VL-BERT & \textbf{77.49} & 55.74 & 71.72 & 39.90 & 34.15 & \textbf{20.10} & 21.17 \\
\bottomrule
\end{tabular}
}
\vspace{-3mm}
\end{table}

\textbf{Experimental Setup.}
We fine-tune all models using the AdamW optimiser~\citep{loshchilov2018decoupled} relying on the same hyper-parameters as in the controlled setup of~\citet{bugliarello-etal-2021-multimodal}.
For few-shot experiments, we instead search three learning rates \{1e-5, 5e-5, 1e-4\} and train for 20 epochs for each \textit{dataset-language-shots} triplet.
Before training, we extract, for each dataset, 36 image regions using a ResNet-101 backbone~\citep{7780459} for mUNITER, xUNITER and UC$^2$; and up to 100 image regions using a ResNeXt-101 backbone~\citep{Xie_2017_CVPR} for M$^3$P---both trained on Visual Genome~\citep{10.1007/s11263-016-0981-7,8578734}.
We train all models on a single NVIDIA V100 (16GB) GPU card.
To increase accessibility to the benchmark by practitioners with limited computing resources: \textbf{1}) we set the number of epochs so that training a baseline can be completed in less than 12 hours per dataset (see \cref{sec:exp-details} for details);\footnote{The only exception is M$^3$P, which is twice as big as the others.} 
\textbf{2}) in addition to reporting performance at every number of shots for few-shot learning, we also define a max-shot evaluation setup to reduce the numbers of runs to one per dataset--language pair.\footnote{Notably, the main bottleneck for current models is the retrieval task as this requires pairing each image with every caption~\citep{Geigle:2022tacl}. In xFlickr\&CO, it takes us $\approx$3 hours per language.} 
For both fine-tuning and few-shot experiments, we evaluate the parameter sets that yield the best validation performance.
For few-shot learning, models are initialised from the English fine-tuned parameter set and further trained on the data in the target language for the given task.

%% file: 05_results.tex
\begin{table}[t]
\caption{Average zero-shot and `translate test' performance of multimodal models trained in English and evaluated on target languages in \benchmark. The best model for each method is in bold.} \label{tab:zero-shot}
\vskip 0.10in
\def\arraystretch{0.93}
\setlength\tabcolsep{2pt}
\small
\center
  \resizebox{0.48\textwidth}{!}{
  \begin{tabular}{lrrcrrrr}
\toprule
\multirow{2}{*}{\small \shortstack{\textbf{Model}}}         &  \multicolumn{1}{c}{\textbf{NLI}}   & \multicolumn{1}{c}{\textbf{QA}} & \multicolumn{1}{c}{\textbf{Reasoning}}  & \multicolumn{4}{c}{\textbf{Retrieval}} \\
  \cmidrule(r){2-2}
  \cmidrule(r){3-3}
  \cmidrule(r){4-4}
  \cmidrule(r){5-8}
 & \multicolumn{1}{c}{XVNLI} & \multicolumn{1}{c}{xGQA}     & \multicolumn{1}{c}{MaRVL} & \multicolumn{2}{c}{xFlickr\&CO}          & \multicolumn{2}{c}{WIT}        \\
 & & & & \multicolumn{1}{c}{IR} & \multicolumn{1}{c}{TR} & \multicolumn{1}{c}{IR} & \multicolumn{1}{c}{TR} \\
\midrule
& \multicolumn{7}{c}{\emph{Zero-Shot}} \\
\midrule

mUNITER & 53.69 & 9.97 & 53.72 & 8.06 & 8.86 & \textbf{9.16} & \textbf{10.48} \\
xUNITER & 58.48 & 21.72 & 54.59 & 14.04 & 13.51 & 8.72 & 9.81 \\
UC$^2$ & \textbf{62.05} & \textbf{29.35} & \textbf{57.28} & \textbf{20.31} & \textbf{17.89} & 7.83 & 9.09 \\
M$^3$P & 58.25 & 28.17 & 56.00 & 12.91 & 11.90 & 8.12 & 9.98 \\
\midrule
& \multicolumn{7}{c}{\emph{Translate test}} \\
\midrule
mUNITER & 73.09 & 49.05 & 63.82 & 40.95 & 36.78 & \textbf{16.25} & 16.64 \\
xUNITER & 72.83 & 49.15 & 64.04 & 36.26 & 30.29 & 13.01 & 14.20 \\
UC$^2$ & 73.67 & 50.19 & 63.09 & 36.03 & 30.37 & 12.70 & 14.11 \\
M$^3$P & 73.37 & 48.83 & 62.52 & 27.74 & 21.29 & 11.53 & 13.63 \\
\midrule
LXMERT & 72.57 & 48.08 & 62.51 & 34.02 & 26.66 & 14.28 & 14.86 \\
UNITER & 73.65 & \textbf{50.62} & 61.92 & 41.04 & \textbf{37.49} & 15.43 & 16.01 \\
ViLBERT & 73.45 & 50.33 & 62.39 & 36.97 & 33.21 & 15.40 & \textbf{16.93} \\
VisualBERT & \textbf{74.12} & 48.72 & 62.35 & \textbf{41.64} & 36.44 & 15.36 & 15.75 \\
VL-BERT & 73.86 & 49.78 & \textbf{64.16} & 38.18 & 31.84 & 15.11 & 16.09 \\
\bottomrule
\end{tabular}
}
\end{table}

\begin{figure*}[t]
 	\centering
 	\subfigure[Wikipedia size (in log scale).]{
 	\includegraphics[width=0.48\textwidth, trim={3cm 0.3cm 4cm 0cm}, clip]{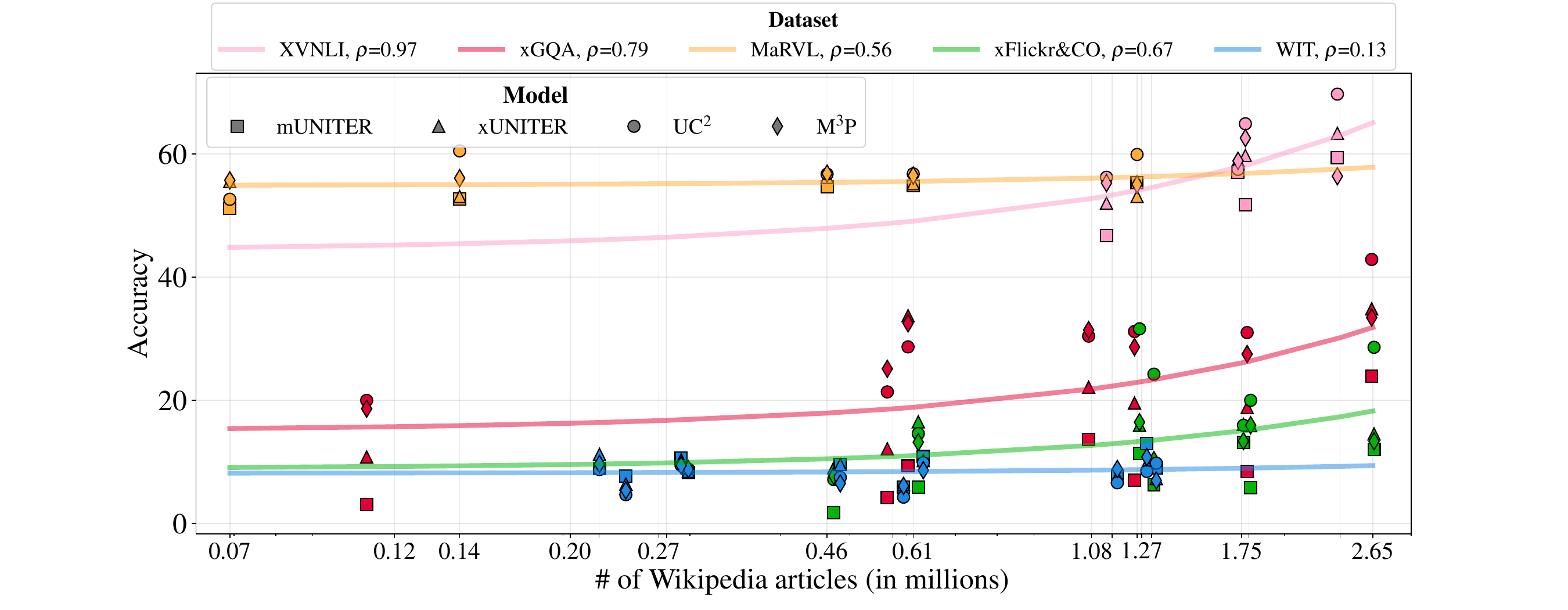}
 	\label{fig:wiki_zero-shot-scores}
 	}%
 	\subfigure[Typological similarity to English.]{
 	\includegraphics[width=0.48\textwidth, trim={3cm 0.3cm 3.5cm 0cm}, clip]{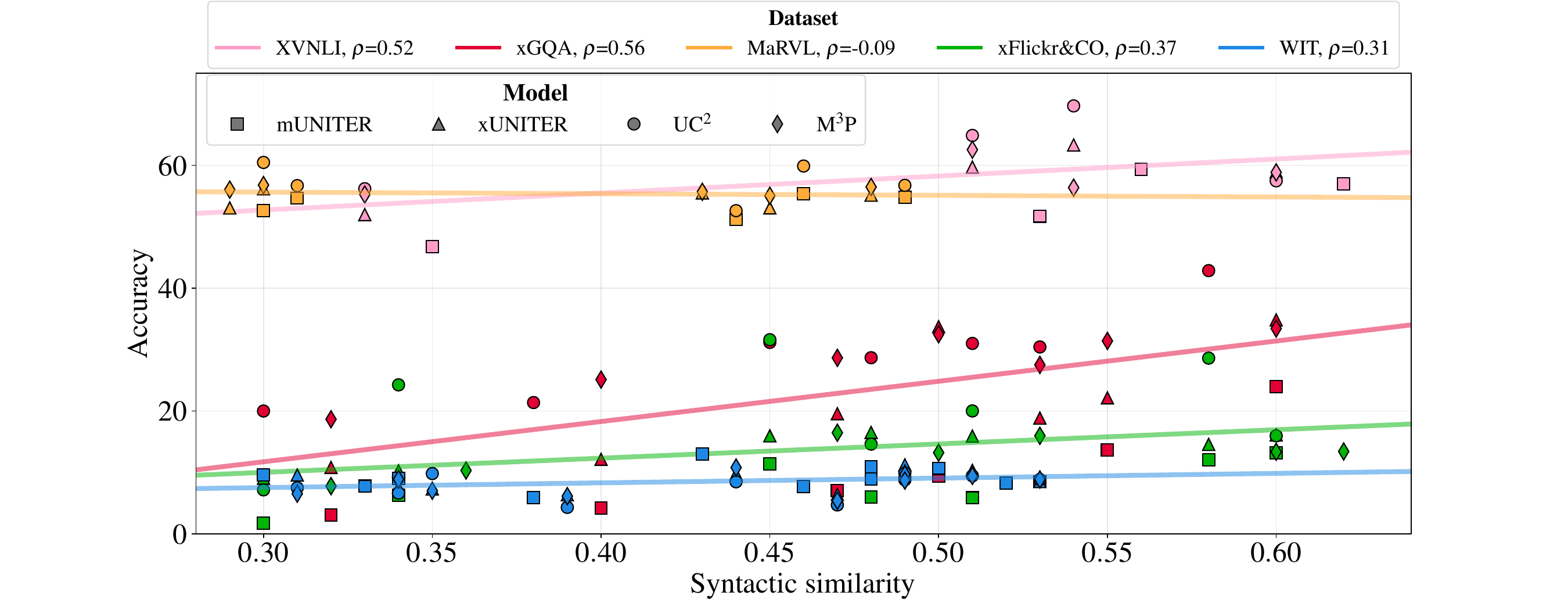}%
 	\label{fig:syntactic_zero-shot-scores}
 	}
 	\vspace{-0.4cm}
 	\caption{Zero-shot performance on individual target languages as a function of some of their properties: Wikipedia size (left) and typological similary to English (right). Pearson's correlation coefficients are also reported in the legends.}
 	\vspace{-0.2cm}
\end{figure*}

\subsection{Zero-shot Learning}

The main zero-shot results, averaged across languages, are reported in \cref{tab:zero-shot}. 
In order to measure the gap between cross-lingual transfer and supervised learning, we also provide the results for the same models trained and evaluated on English test data in \cref{tab:english}. 
The performance metric is accuracy for all tasks except cross-modal retrieval, which uses Recall@1.\footnote{Note that these metric are identical whenever image--caption pairs are unique. This is the case for all of xFlickr\&CO data points and for more than 90\% of the samples in WIT (\cref{sec:WIT}).} 
Finally, for completeness, the full results subdivided by language are detailed across \cref{sec:perlanguage}. 
Next, we compare these scores across different dimensions.

\rparagraph{Transfer Method}
The results in \cref{tab:zero-shot} clearly demonstrate the superiority of `translate test' transfer over zero-shot model transfer via multilingual encoders. 
The difference between the best models for each method reaches 12.1 points for XVNLI, 21.6 for xGQA, 6.9 for MaRVL, 21.3 for xFlickr\&CO IR, 19.6 for xFlickr\&CO TR, 7.1 for WIT IR, and 6.5 for WIT TR. Thus, gains are especially remarkable in cross-modal retrieval tasks and grounded question answering, where the performance is nearly doubled by virtue of `translate test' transfer.

\rparagraph{Multilingual Models} 
Even within each transfer method, we observe considerable variance among individual multilingual multimodal models. 
For instance, for zero-shot model transfer via multilingual encoders, UC$^2$ is consistently better across the board (except for WIT) and surpasses the other models by a large margin, especially in German, French, Japanese and Mandarin---languages in which the authors had translated CC and pretrained UC$^2$ on (\cref{sec:perlanguage}). 
Remarkably, this tendency is less accentuated in MaRVL, where pretraining Mandarin multimodal data is still insufficient to tackle the out-of-distribution nature of MaRVL's culture-specific concepts~(\cref{tab:marvl} in~\cref{sec:perlanguage}).
Nonetheless, these results prove how the simple `translate pretrain' approach of UC$^2$ can be an effective baseline for multilingual transfer in multimodal pretrained models.
As for `translate test' transfer, no clear winner emerges across the monolingual models, whereas mUNITER generally performs better among multilingual models limited to cross-modal retrieval tasks. 
We also note that massively multilingual models (top rows) are often on a par with their monolingual counterparts (bottom rows).

\rparagraph{Transfer Gap}
Comparing the results on the English test set (\cref{tab:english}) with those averaged across the other languages (\cref{tab:zero-shot}) reveals an extremely large gap in performance due to cross-lingual transfer. 
Considering the best multilingual encoders for each task, the gap is 14.8 points for XVNLI, 26.5 for xGQA, 14.6 for MaRVL, 24.2/23.0 for xFlickr\&CO IR/TR, and 10.7/11.9 for WIT IR/TR. 

\rparagraph{Task Complexity}
There emerge stark contrasts among the model performances in each individual task. Nonetheless, we remind the reader that the number of classes to predict---and hence random baselines---vary across tasks: XVNLI 3, xGQA 1{,}842, and MaRVL 2. 
Moreover, some tasks include a collection of languages that are more homogeneous (e.g.\ XVNLI contains mostly Indo-European languages) whereas others are more diverse, such as WIT. 
Performance is expected to suffer the most in the latter group. Moreover, focusing on cross-modal retrieval tasks, scores are higher in xFlickr\&Co than in WIT. 
This is arguably due to the fact that, while all models were pretrained on an image--text matching task, WIT captions are distributionally distant from standard captions as they mostly describe entities.

\begin{figure*}[t]
 	\centering
 	\includegraphics[width=\linewidth, trim={0cm 0cm 0cm 0cm}, clip]{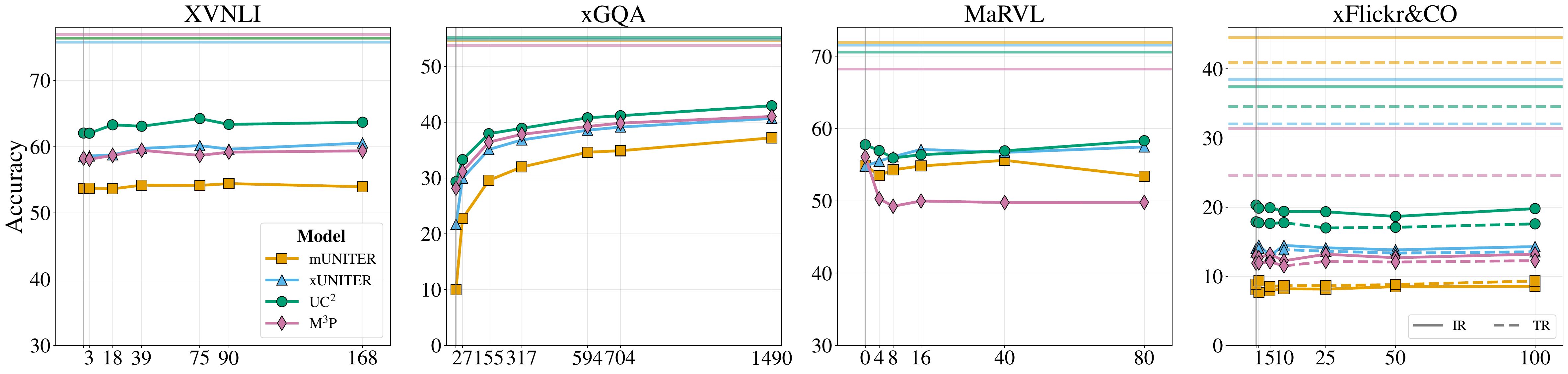}%
 	\vspace{-0.2cm}
 	\caption{Few-shot performance on \benchmark tasks (cross-lingual average) against different data sizes for target language fine-tuning. Values in MaRVL are only averaged across the 3 languages with available few-shot data. Horizontal lines report English performance.}
 	\label{fig:fewshot-curves-avgV2}
 	\vspace{-0.1cm}
\end{figure*}
\begin{figure*}[t]
 	\centering
 	\includegraphics[width=\linewidth, trim={0cm 0cm 0cm 0cm}, clip]{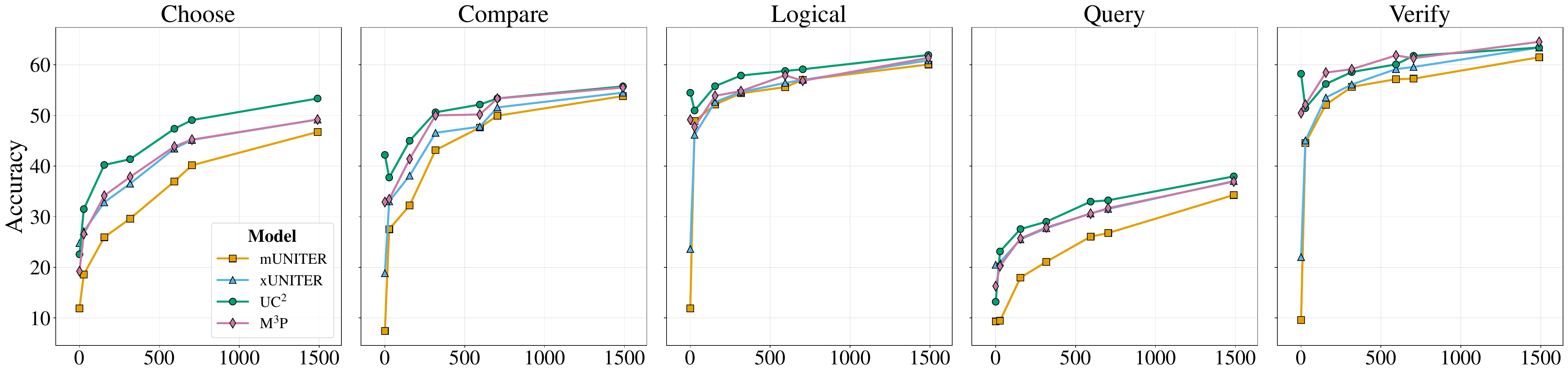}%
 	\vspace{-0.2cm}
 	\caption{Analysis of few-shot results on xGQA over five structurally different questions types.}
 	\label{fig:fewshot-xgqa-struct}
 	\vspace{-0.3cm}
\end{figure*}

\rparagraph{Explanatory Variables}
Finally, we assess the impact of a series of explanatory factors on the performance in each individual target language.
More specifically, we estimate its correlation \textbf{1}) with the amount of each language's unlabelled textual data available for pretraining in terms of Wikipedia articles (as of January 2022) in	\cref{fig:wiki_zero-shot-scores}; and \textbf{2}) with the typological similarity between the source (English) and the target language in cross-lingual transfer in \cref{fig:syntactic_zero-shot-scores}. 
The latter is calculated as the cosine similarity between vectors of morpho-syntactic features extracted from WALS \citep{wals} by \citet{littell-etal-2017-uriel}.\footnote{\url{https://github.com/antonisa/lang2vec}.} 
According to the received wisdom from text-only cross-lingual transfer \citep{ponti-etal-2020-xcopa,ruder-etal-2021-xtreme}, both these factor usually play a pivotal role in determining the downstream performance of each target language. 
While we find this to partially hold true for Wikipedia size (with the possible exception of WIT), we find that correlations of typological similarity are weaker (and even negative for MaRVL), implying that models may struggle comparatively on many target languages despite different similarities to English.

\subsection{Few-shot Learning}
As we experimented with a range of data sizes for $k$-shot fine-tuning in the target language, we \textbf{1}) plot these values against the corresponding performance in \cref{fig:fewshot-curves-avgV2} and \textbf{2}) also report the exact scores for the maximum $k$ in \cref{tab:few-shot-max}, which acts as an alternative benchmark for low-compute practitioners to evaluate few-shot learning.
These results aid in determining the sample efficiency of the models, i.e., how much they benefit from observing each new data point in the target language. 
In addition, \cref{tab:few-shot-auc} (\cref{sec:more-few-results})
reports the Area Under the Curve (AUC) for accuracy across different $k$ values, which instead measures how consistently well a model performs independently from data size.

\begin{table}[t]
\caption{Max-shot performance (with the largest $k$) averaged across languages with few-shot training data.} 
\label{tab:few-shot-max}
\setlength\tabcolsep{2pt}
\small
\center
  \begin{tabular}{lrrcrr}
\toprule
\multirow{2}{*}{\small \shortstack{\textbf{Model}}}         &  \multicolumn{1}{c}{\textbf{NLI}}   & \multicolumn{1}{c}{\textbf{QA}} & \multicolumn{1}{c}{\textbf{Reasoning}}  & \multicolumn{2}{c}{\textbf{Retrieval}} \\
  \cmidrule(r){2-2}
  \cmidrule(r){3-3}
  \cmidrule(r){4-4}
  \cmidrule(r){5-6}
 & \multicolumn{1}{c}{XVNLI} & \multicolumn{1}{c}{xGQA}     & \multicolumn{1}{c}{MaRVL} & \multicolumn{2}{c}{xFlickr\&CO}    \\
 & & & & \multicolumn{1}{c}{IR} & \multicolumn{1}{c}{TR} \\
\midrule
mUNITER & 53.95 & 37.21 & 53.41 & 8.54 & 9.32 \\
xUNITER & 60.55 & 40.68 & 57.46 & 14.30 & 13.54 \\
UC$^2$ & \textbf{63.68} & \textbf{42.95} & \textbf{58.32} & \textbf{19.79} & \textbf{17.59} \\
M$^3$P & 59.36 & 41.04 & 49.79 & 13.21 & 12.26 \\
\bottomrule
\end{tabular}
\vspace{-2mm}
\end{table}

\rparagraph{Gains from Few-shot Learning}
Contrary to text-only multilingual tasks, where even few examples in the target language are sufficient to significantly improve the model performance \citep{lauscher-etal-2020-zero}, we find that this is largely not the case in multimodal settings. 
\cref{fig:fewshot-curves-avgV2} reveals that learning curves are flat for XVNLI, MaRVL, and xFlickr\&CO. Moreover, inspecting the gap between zero-shot and max-shot performance in \cref{tab:few-shot-diffs} (\cref{sec:more-few-results}), gains from few-shot learning are consistent but modest. 

A notable exception is xGQA, where as few as 27 examples increase performance dramatically and accuracy continues to grow with the sample size.
This difference is possibly due to the highly-structured nature of xGQA, where texts are generated from templates and are extremely short, and its larger label space. 
To investigate the latter, we follow~\citet{pfeiffer2021xgqa} and evaluate the accuracy for each structurally different question type. 
\cref{fig:fewshot-xgqa-struct} shows that increasing the number of examples has the least impact on \emph{Verify}-type questions, which consist of (binary) answers/labels (\emph{\{Yes, No\}}). 
These results are in line with our few-shot experiments on the other tasks, where few-shot tuning displays negligible impact on cross-lingual transfer when the label space is small. 
The largest impact in xGQA is observed on \emph{Choose}-type and \emph{Query}-type questions, where the diversity of the label space is large. 
This again shows that the multilingual misalignment of the prediction space negatively impacts zero-shot results, which can be mitigated by fine-tuning even on a few examples in the target language.

\begin{figure}[t]
 	\centering
 	\includegraphics[width=\linewidth, trim={0cm 0cm 0cm 0cm}, clip]{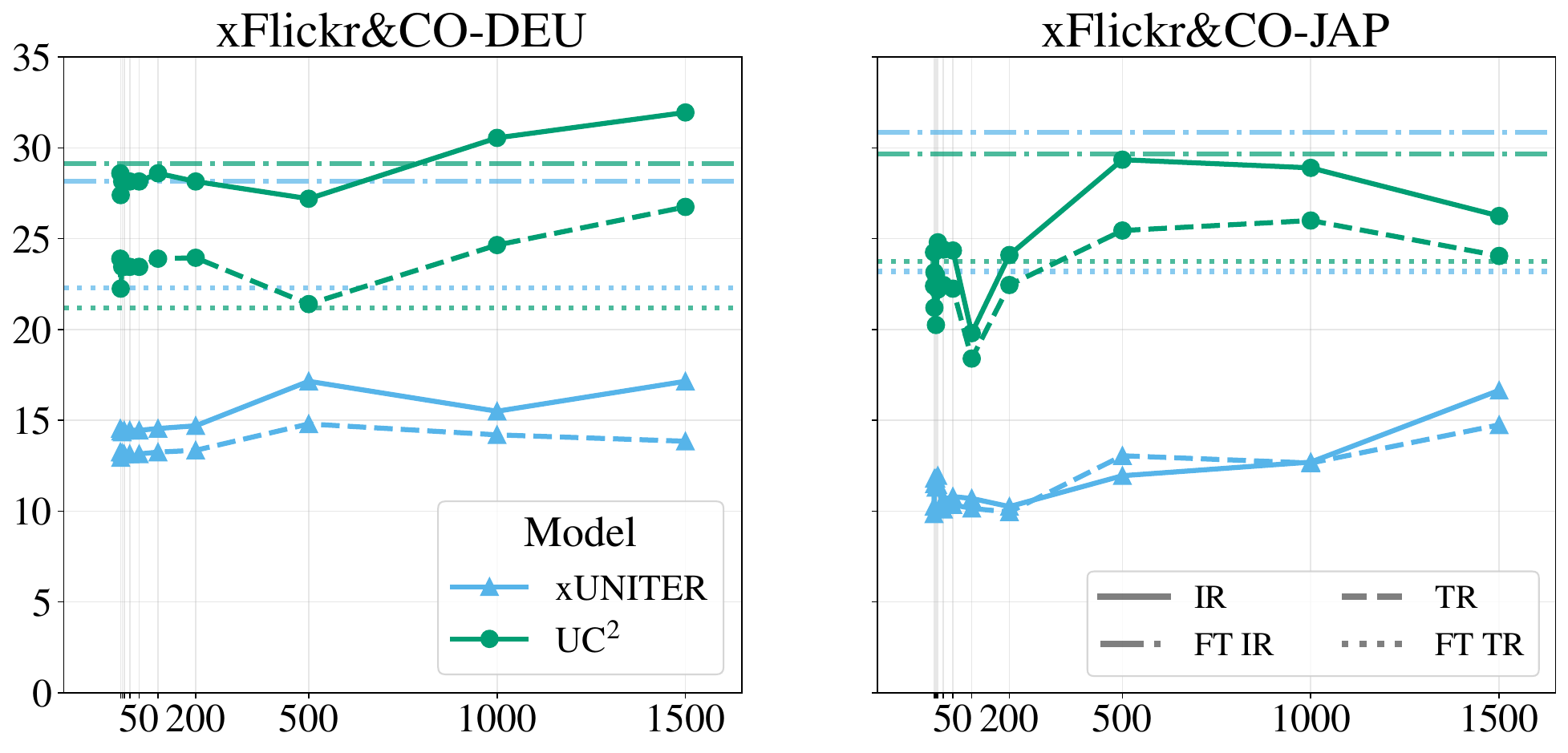}%
 	\vspace{-0.2cm}
 	\caption{Retrieval accuracy on \german{} and \japanese{} xFlickr\&CO splits with target language fine-tuning (horizontal lines) and more shots.}
 	\label{fig:fewshot-curves-xflickr-deja}
 	\vspace{-0.25cm}
\end{figure}

On the other hand, current models may require many more examples than the maximum of $k$-shot we established for the other tasks. 
We test this hypothesis in a few-shot learning experiment with more examples in \textsc{jpn} and \textsc{deu} on xFlickr\&CO.
The results in \cref{fig:fewshot-curves-xflickr-deja} demonstrate that examples in the order of the thousands may be needed to achieve significant gains. 
This leaves open an important challenge to make future models more sample efficient, in more resource-poor settings like those considered here.

\textbf{Multilingual Encoders.} 
Again, based on \cref{tab:few-shot-max} the model that yields the highest scores is UC$^2$ in all tasks. However, we also note that some models are particularly brittle,and few-shot adaptation can lead them to revert to chance-level performance, as is the case of M$^3$P when fine-tuning on MaRVL's target languages \indonesian, \chinese, and \turkish.

%% file: 07_conclusion.tex
We introduced the Image-Grounded Language  Understanding Evaluation (IGLUE) benchmark, aiming to facilitate modelling and evaluation in multilingual vision-and-language research. 
IGLUE brings together 5 datasets across 4 structurally different tasks in 20 diverse languages, and provides standardised data splits and evaluation protocols to train and evaluate in zero-shot and few-shot transfer setups. 

We set strong baselines and carry out analyses that shed new light into multilingual \vl models, now enabled by the creation of the IGLUE benchmark. In particular, we find that \textbf{1}) `translate test' transfer vastly outperforms zero-shot transfer via multilingual encoders; \textbf{2}) few-shot learning requires thousand of examples before yielding gains; \textbf{3}) unlabelled textual data size and typological distance from English are weaker predictors of models performance than usually observed in text-only tasks; 
\textbf{4}) results vary significantly across tasks, with multicultural visual reasoning and cross-modal retrieval having the largest transfer gaps.
Given the diversity and manifoldness of each task, we explicitly set IGLUE as a multi-faceted benchmark, without a single average score.

Along with putting forward an evaluation suite for current and future multilingual \vl models, IGLUE also opens up new opportunities for future research that goes beyond the scope of this paper: such as analysing single-source versus multi-source transfer for \vl tasks, conducting multi-task learning, or investigating staged multi-task transfer. 
A longer-term outlook concerns including other tasks and languages into the benchmark, as well as reaching beyond the image-only visual modality, towards multimodal tasks that rely on processing videos~\cite{li2021value} and speech~\cite{Yang2021SUPERBSP} in different languages.

%% file: 08_app.tex
\section{Datasets}\label{sec:datasets}
In this section, we provide more details about the datasets and splits that we introduce in \benchmark, focusing on reproducibility of our experiments and any follow-up work that leverages the IGLUE benchmark.
We also report how few-shot samples were collected for each dataset. 
In particular, we use the notion of `annotation context' to define what a `shot' means in each dataset (see details below).
\cref{tab:few-shot} lists the size of few-shot splits across all datasets.

\subsection{XVNLI}

\begin{figure}[htp]
 	\centering
 	\subfigure[Uniform distribution of labels across splits.]{
 	\includegraphics[width=0.49\textwidth, trim={1.8cm 0cm 2cm 0cm}, clip]{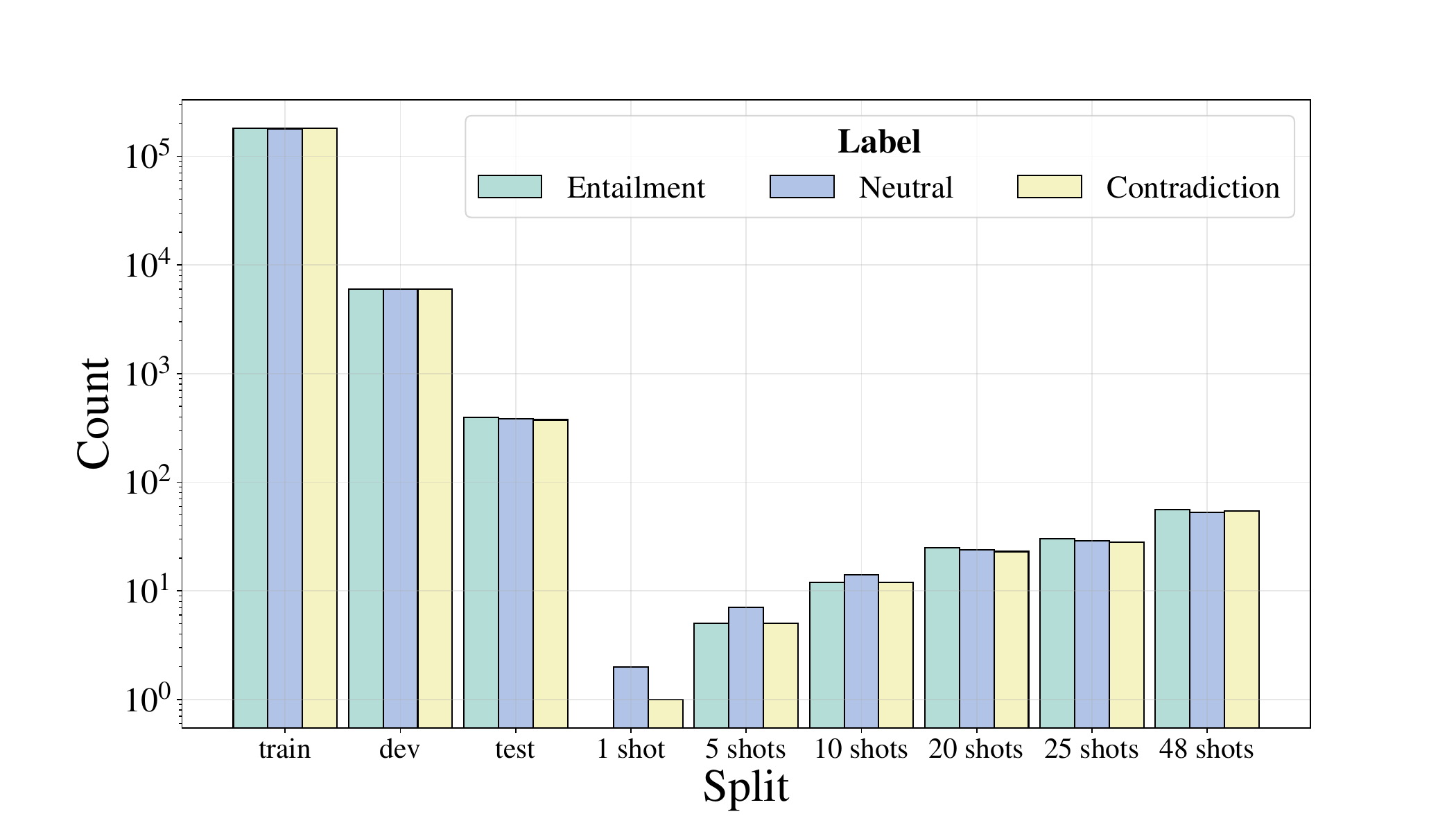}
 	\label{fig:xvnli-labels}
 	}
 	\subfigure[Similar test sentence length distribution across languages.]{
 	\includegraphics[width=0.49\textwidth, trim={1cm 0cm 2cm 0cm}, clip]{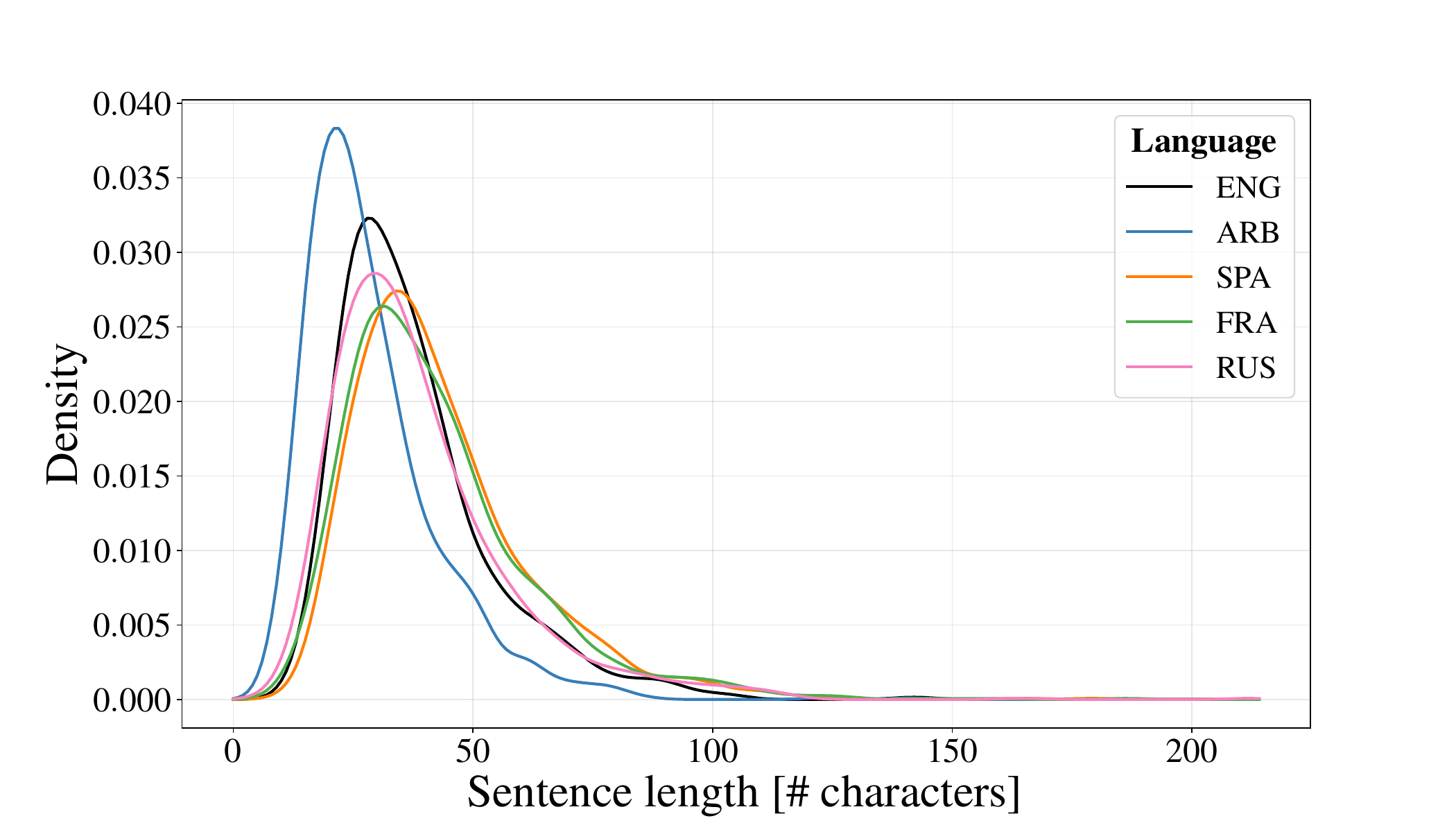}%
 	\label{fig:xvnli-lens}
 	}
 	\caption{Statistical properties of XVNLI.} \label{fig:xvnli-stats}
\end{figure}

\cref{fig:xvnli-stats} illustrates statistical properties of XVNLI. 
In~\cref{fig:xvnli-labels}, we see that the distribution of labels is uniform across splits. 
\cref{fig:xvnli-lens} instead shows that the distribution of sentence lengths---in number of characters---in the test split is similar across languages, and often shorter for \arab.
\vspace{-0.3cm}

\textbf{Few-shot Data.}
We extract few-shot samples from the original SNLI test split, which was translated by~\citet{agic-schluter-2018-baselines}.
This was done prior to defining our test split, and by sampling all the examples corresponding to a given image, so as to avoid data leakage between training and evaluation data.
In XVNLI, multiple samples (hypotheses and answers) are associated to a given image (premise).
We hence consider an image as a single shot, as it serves as the context for an annotator to create multiple samples.
Following~\citet{pfeiffer2021xgqa}, we randomly sample up to 48 images as contexts, resulting in up to 168 hypotheses and answers.

\subsection{xGQA}
\textbf{Few-shot Data.}
We reuse the few-shot splits defined by~\citet{pfeiffer2021xgqa}, which also include development data in the target languages. The authors sampled 48 images, giving up to 1{,}490 questions translated for each language.

\begin{table}[t]
\caption{Few-shot statistics. * for MaRVL, we collected few-shot data in three of the five languages due to challenges in annotations.}
\label{tab:few-shot}
\vskip 0.10in
\setlength\tabcolsep{3pt}
\small
\center
  \begin{tabular}{llrrrrrr}
\toprule
\textbf{Dataset} & \textbf{Metric} & & & & & & \\
\midrule
\multirow{3}{*}{\small \shortstack{XVNLI}} & \# shots & 1 & 5 & 10 & 20 & 25 & 48 \\
& \# images & 1 & 5 & 10 & 20 & 25 & 48 \\
& \# samples & 3 & 18 & 39 & 75 & 90 & 168 \\
\midrule
\multirow{3}{*}{\small \shortstack{xGQA}} & \# shots & 1 & 5 & 10 & 20 & 25 & 48 \\
& \# images & 1 & 5 & 10 & 20 & 25 & 48 \\
& \# samples & 27 & 155 & 317 & 594 & 704 & 1{,}490 \\
\midrule
\multirow{3}{*}{\small \shortstack{MaRVL*}} & \# shots & 1 & 2 & 4 & 10 & 10$\times$2 & \\
& \# images & 8 & 16 & 32 & 80 & 80 & \\
& \# samples & 4 & 8 & 16 & 40 & 80 & \\
\midrule
\multirow{3}{*}{\small \shortstack{xFlickr\&CO}} & \# shots & 1 & 5 & 10 & 25 & 50 & 100 \\
& \# images & 1 & 5 & 10 & 25 & 50 & 100 \\
& \# samples & 1 & 5 & 10 & 25 & 50 & 100 \\
\bottomrule
\end{tabular}
\end{table}

\subsection{MaRVL}
\rparagraph{Few-shot Data}
We follow the annotation protocol of~\citet{liu-etal-2021-visually} to collect few-shot samples. 
For each language, we first ask native speakers to select 10 \emph{concepts} among the ones identified by the authors in order to collect in-domain training data.
The annotators are then required to retrieve at least 8 images per concept. 
When doing so, we further provide the MaRVL test images and require them to select different ones to avoid any visual leakage betweem few-shot and test samples.
We finally sample 8 images per concept, randomly pair them and ask annotators to write a caption that is true for two pairs and false for the other two pairs.
These 8 images then constitute an `annotation context' and as such we consider the resulting data as one shot (i.e. one shot, eight images and one sentence giving four data points).
As image collection is the most time-consuming step in the pipeline, we expand our few-shot data by re-shuffling and re-annotating each image pair (`$10\times2$' setup), resulting in 80 data points per language.
Notably, due to the challenges in collecting few-shot data in low-resource languages (i.e. Swahili and Tamil), we only provide few-shot samples for three languages (i.e. Indonesian, Mandarin and Turkish).

\subsection{xFlickr\&CO} \label{sec:xflickrco}
\begin{figure}[htp]
 	\centering
 	\subfigure[Flickr30K-like instructions to annotate xFlickr\&CO data.]{
 	\frame{\includegraphics[width=0.48\textwidth, trim={2.5cm 1cm 4cm 1cm}, clip]{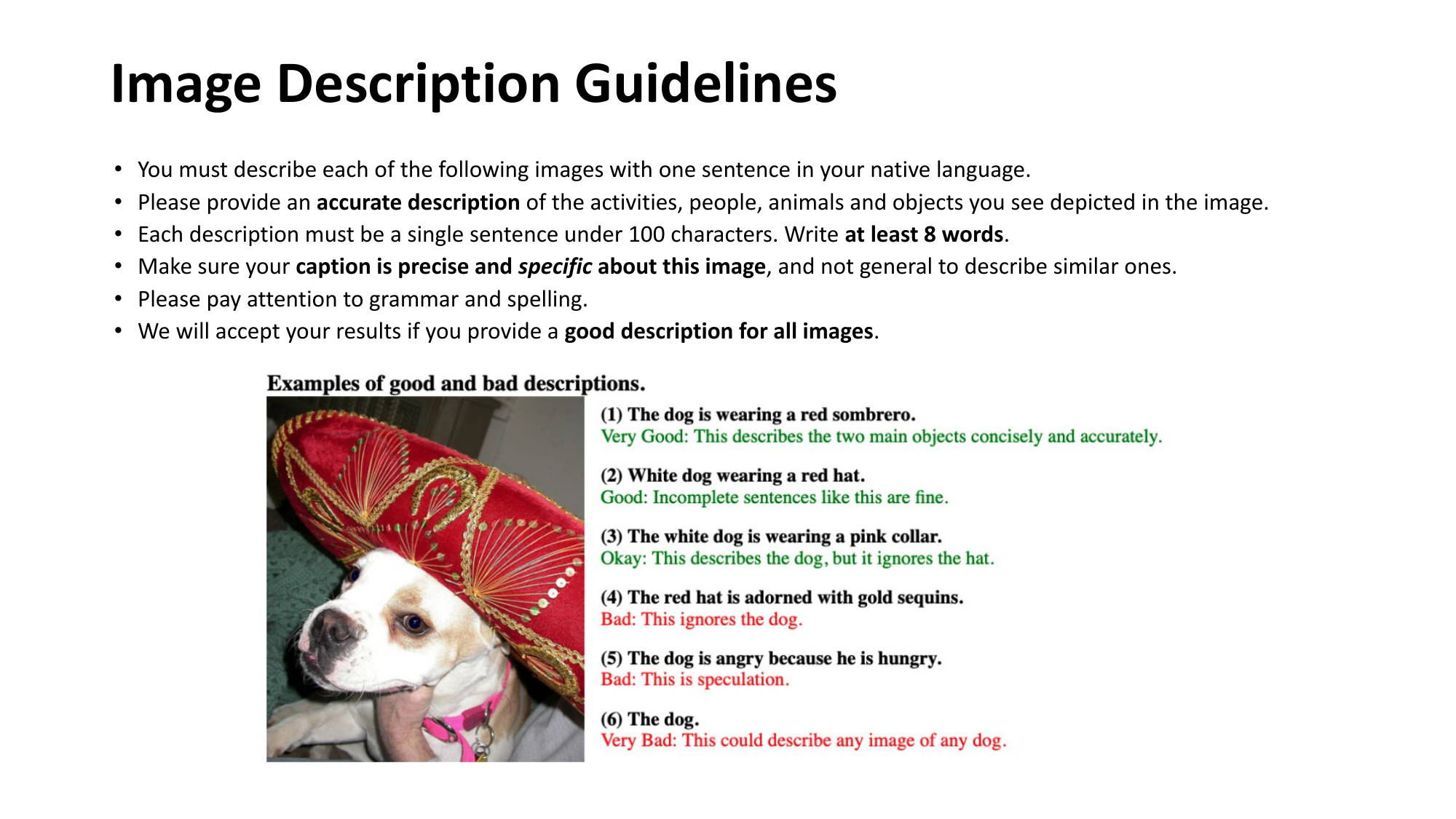}}%
 	\label{fig:xflickrco-instructions}
 	}
 	\subfigure[Additional Flickr30K examples used to further guide annotators.]{
 	\frame{\includegraphics[width=0.48\textwidth, trim={2cm 0.5cm 1.5cm 1cm}, clip]{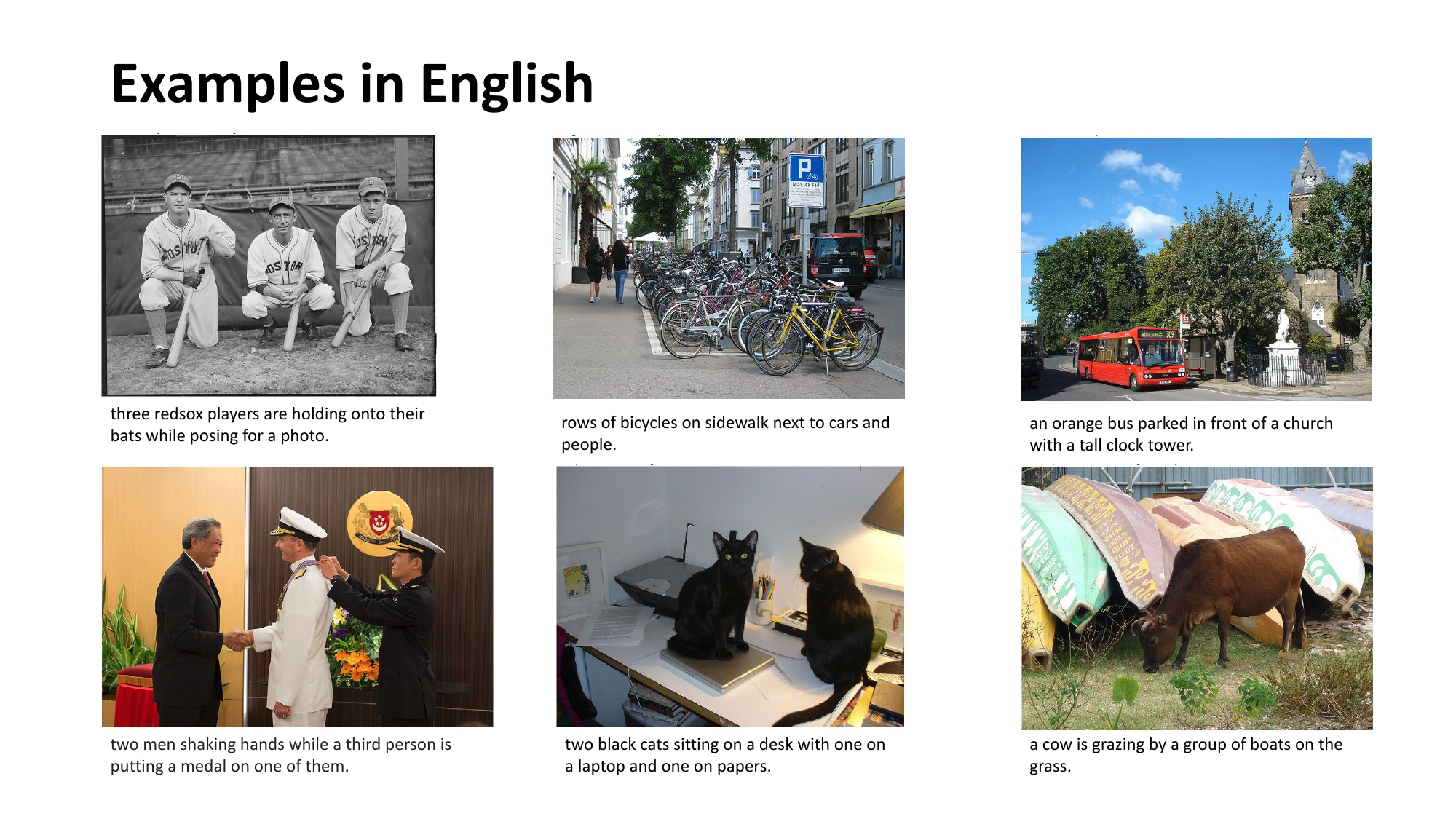}}%
 	\label{fig:xflickrco-examples}
 	}
 	\subfigure[Instructions given to native speakers to validate the captions.]{
 	\frame{\includegraphics[width=0.48\textwidth, trim={2cm 4cm 3cm 1cm}, clip]{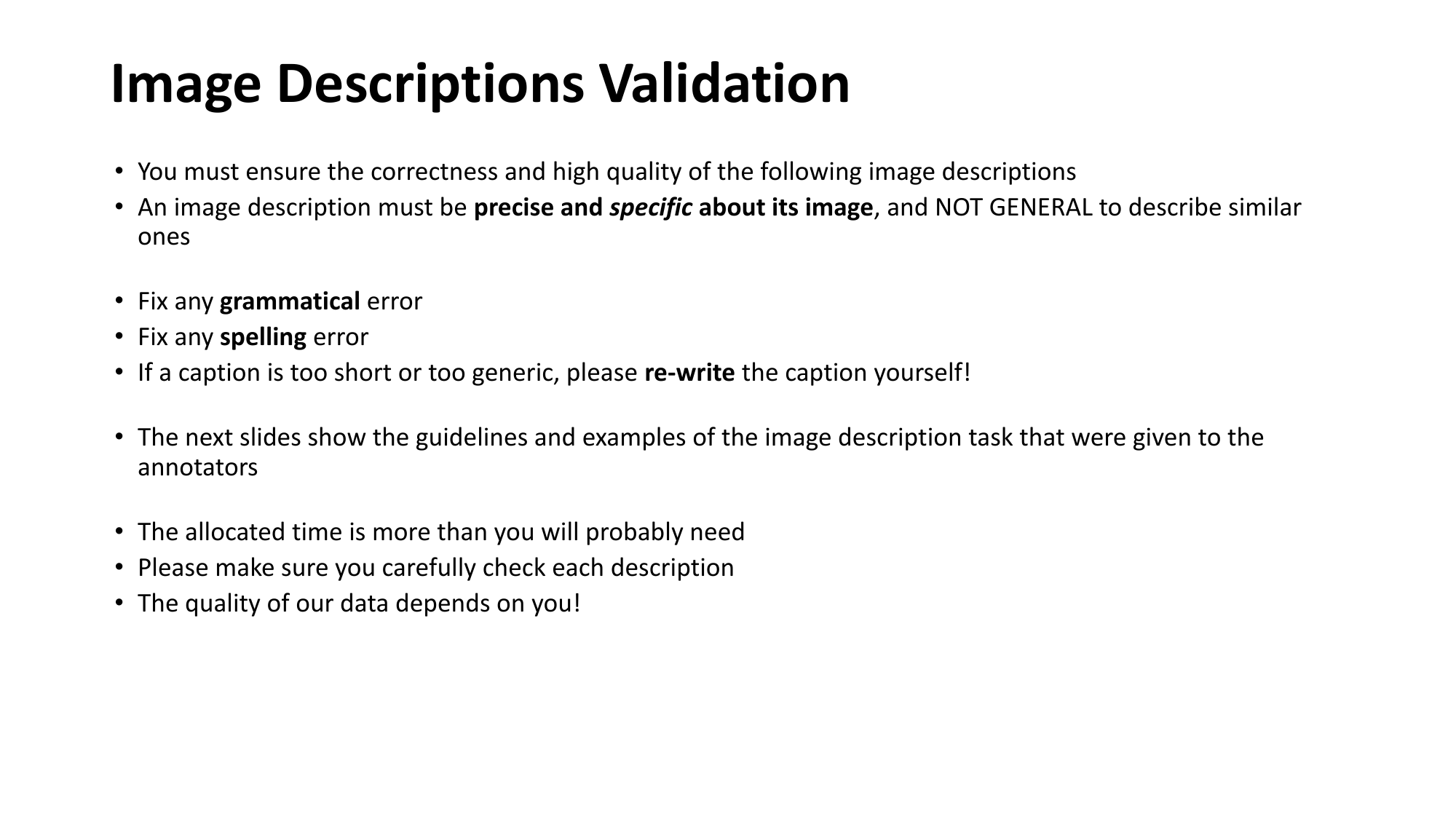}}%
 	\label{fig:xflickrco-validation}
 	}
 	\caption{Annotation guidelines for xFlickr\&CO.} 
 	\label{fig:xflickrco-guidelines}
\end{figure}

We follow the Flickr30K guidelines to collect our multilingual annotations, as shown in \cref{fig:xflickrco-instructions}, which we supplement with examples from Flickr30K (\cref{fig:xflickrco-examples}).
In particular, we limit captions length based on previous work: 40 characters for \chinese, 140 for \turkish{} and 100 for the rest.
\cref{fig:xflickrco-validation} shows the additional guidelines provided to validators.
The resulting corpora were finally verified by native speakers known by the authors.
Notably, the same guidelines were used to annotate both Flickr30K and COCO images.
\cref{fig:xflickrco-guidelines} shows the distribution of caption lengths.

\begin{figure}[t]
 	\centering
 	\includegraphics[width=0.45\textwidth, trim={1cm 0.4cm 2cm 0cm}, clip]{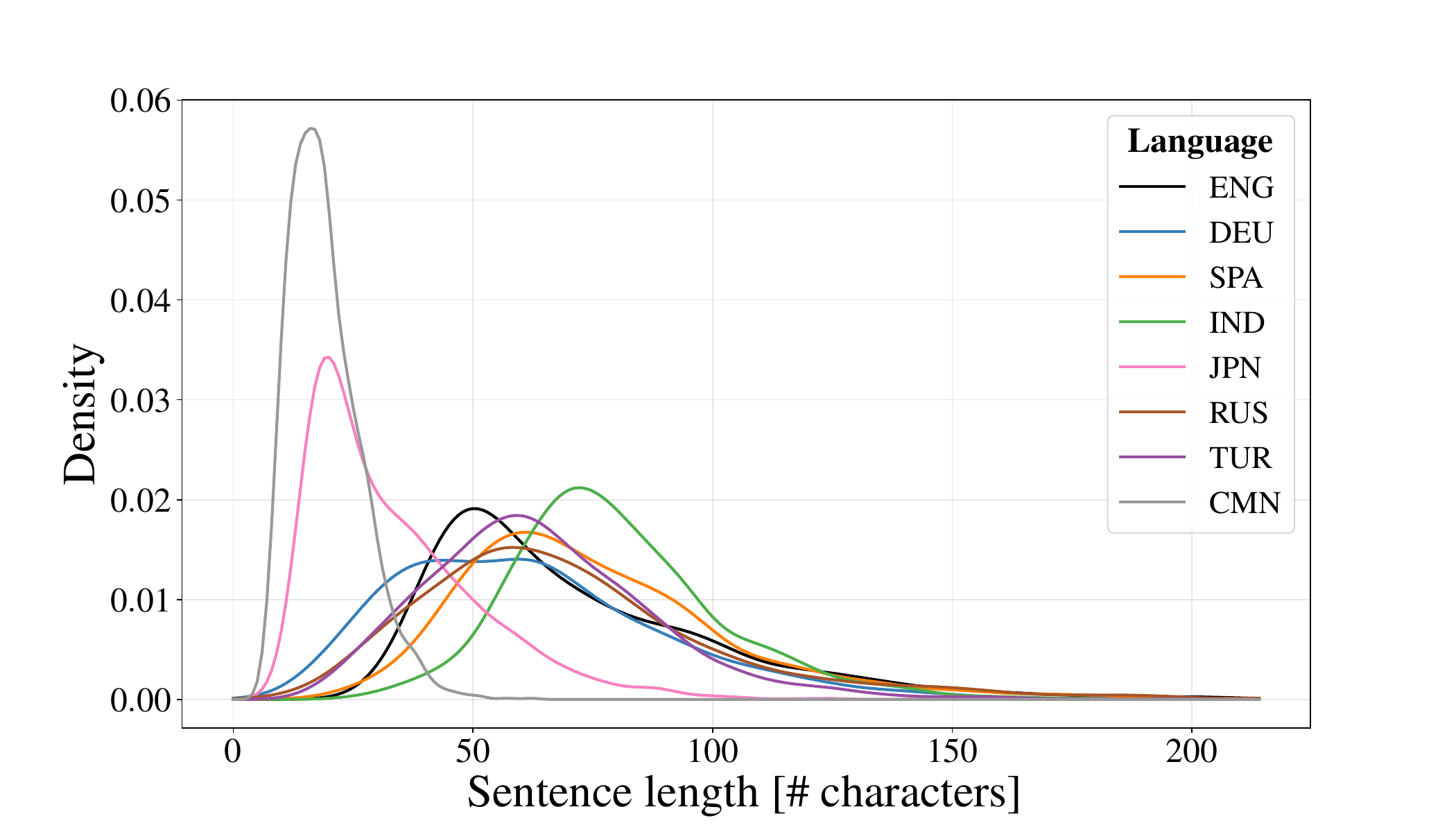}
 	\vspace{-0.2cm}
 	\caption{Test sentence length distribution in xFlickr\&CO.} \label{fig:xflickrco-lens}
 	\vspace{-0.2cm}
\end{figure}

\rparagraph{Few-shot Data}
We use the same guidelines illustrated in~\cref{fig:xflickrco-guidelines} to annotate few-shot image descriptions from 100 images samples from the Flickr30K training set.
German captions are extracted from Multi30K~\citep{elliott-etal-2016-multi30k}, Mandarin ones from Flickr30K-cn~\citep{10.1145/3432246}, and Turkish ones from TasvirEt~\citep{unal2016tasviret} -- all independently collected from the English captions.

\subsection{WIT} \label{sec:WIT}
\begin{figure}[t]
 	\centering
 	\includegraphics[width=0.4\textwidth, trim={0cm 0cm 0.5cm 0cm}, clip]{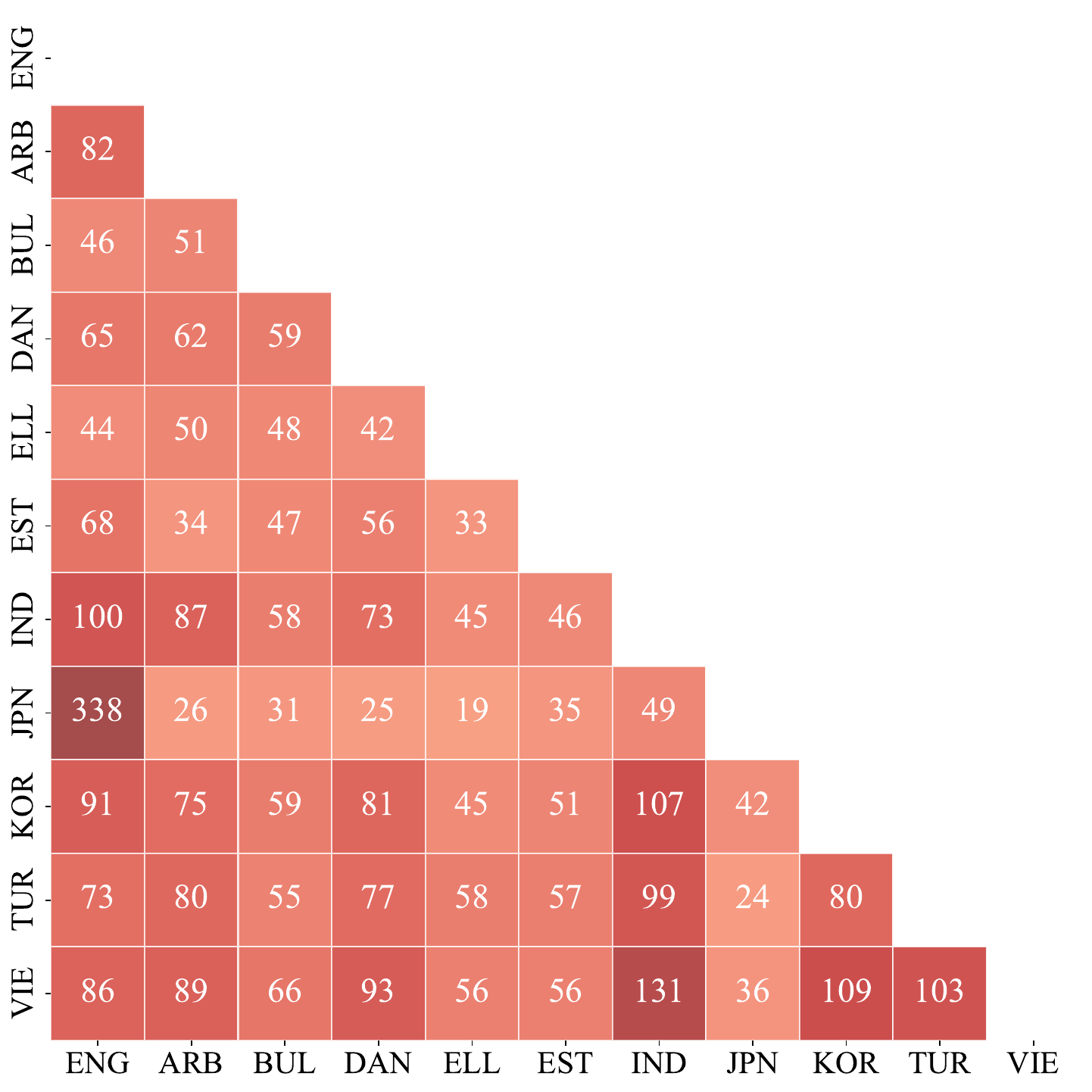}
 	\vspace{-0.2cm}
 	\caption{Number of images overlapping between any two languages in the WIT test splits.} 
 	\label{fig:wit-overlap}
\end{figure}

\begin{figure}[t]
 	\centering
 	\includegraphics[width=0.45\textwidth, trim={1cm 0cm 3.5cm 2cm}, clip]{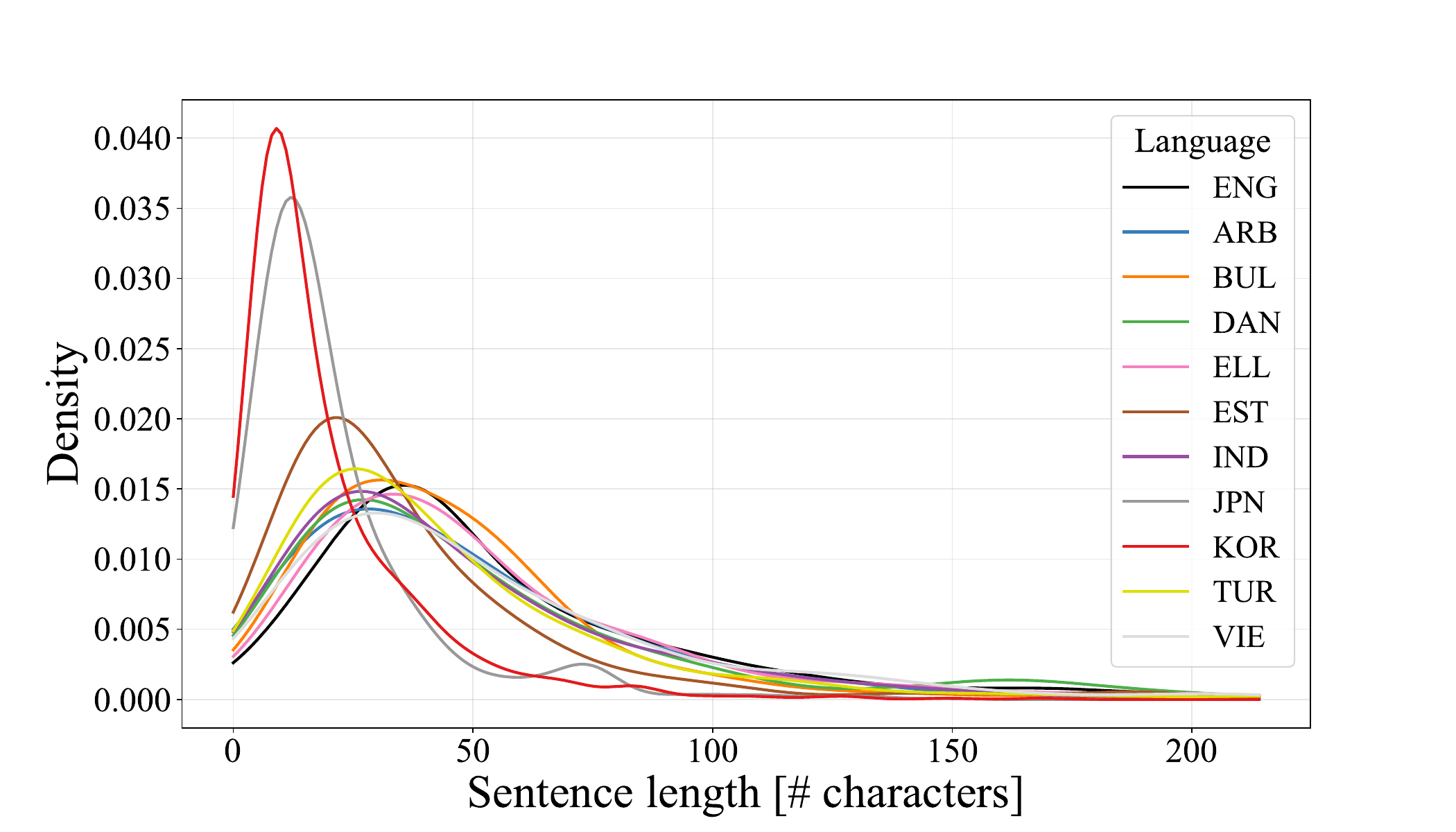}
 	\vspace{-0.2cm}
 	\caption{Test sentence length distribution in WIT.}
 	\label{fig:wit-lens}
\end{figure}

We extract 500{,}000 image captions from the \texttt{caption\_reference\_description} field of the English portion of the WIT dataset. 
This field corresponds to the captions shown on the Wikipedia pages directly below the images, and it tends to be the most topical and relevant description~\citep{10.1145/3404835.3463257}.
In particular, we sample among the images that were released by the Wikimedia Foundation for the corresponding Kaggle challenge,\footnote{\url{https://www.kaggle.com/c/wikipedia-image-caption/data}.} which do not include any identifiable personal information.
This challenge also provides test data that was extracted from WIT prior to its public release.\footnote{While we find no overlap between train and test image URLs, we find that a few \texttt{base64} image encodings match between train and test splits: 1 image for \arab, \english{} and \japanese; 2 images for \bulgarian, \korean{} and \turkish; 3 images for \indonesian{} and \vietnamese; 0 for the other languages.} 
We use these data to define our multilingual test splits in order to let future practitioners evaluate models that were also trained on the WIT data.
Per-language statistics are listed in~\cref{tab:wit-stats}. 
We note that a given image is mapped to a single caption more than 90\% of the times in every language except for English (where this happens around 70\% of the times and which is not used as an evaluation language in \benchmark).
\cref{fig:wit-overlap} shows the overlap between test images across languages, and \cref{fig:wit-lens} illustrates the distribution of caption lengths across languages in our WIT test splits.

\textbf{Few-shot Data.}
Due to the demanding computational resources to evaluate image--text retrieval systems, we only provide few-shot splits for this task in xFlickr\&CO.

\begin{table}[H]
\caption{WIT test data statistics. \% unique measures the percentage of images that only have a single caption associated with them.}
\label{tab:wit-stats}
\vskip 0.10in
\setlength\tabcolsep{3pt}
\small
\center
  \resizebox{0.48\textwidth}{!}{
  \begin{tabular}{lrrrrrrrrrrr}
\toprule
\textbf{} & \multicolumn{10}{c}{\textbf{Language}} \\
\textbf{Metric} & \multicolumn{1}{c}{\colorbox{gray!20}{\textbf{\english}}} & \multicolumn{1}{c}{\textbf{\arab}} & \multicolumn{1}{c}{\textbf{\bulgarian}} & \multicolumn{1}{c}{\textbf{\danish}} & \multicolumn{1}{c}{\textbf{\greek}} & \multicolumn{1}{c}{\textbf{\estonian}} & \multicolumn{1}{c}{\textbf{\indonesian}} & \multicolumn{1}{c}{\textbf{\japanese}} & \multicolumn{1}{c}{\textbf{\korean}} & \multicolumn{1}{c}{\textbf{\turkish}} & \multicolumn{1}{c}{\textbf{\vietnamese}} \\
\midrule
\# images & \colorbox{gray!20}{~~685} & 792 & 806 & 814 & 541 & 780 & 854 & 842 & 889 & 681 & 869 \\
\# samples & \colorbox{gray!20}{1000} & 890 & 860 & 891 & 570 & 874 & 901 & 1000 & 931 & 721 & 946 \\
\% unique & \colorbox{gray!20}{~70.9} & 91.2 & 94.9 & 93.0 & 97.2 & 93.6 & 96.0 & 90.8 & 95.8 & 95.6 & 93.7 \\
\bottomrule
\end{tabular}
}
\end{table}

\subsection{Ethics Statement}
For MaRVL and xFlickr\&CO, we collected data from native speakers spread across the world in languages that are under-represented in current vision and language datasets.
Our efforts were motivated by the belief that vision and language tasks need to better reflect the linguistic (and cultural) variety present in the world.
We hired workers from the crowdsourcing platform \url{prolific.co}. 
The annotators' identities were anonymous to us, and each of them was paid £10-20/hour (even if their data were not ultimately used).

\begin{table}[!t]
    \caption{Pretrained multilingual \vl encoders.}
    \label{tab:models}
    \vskip 0.05in
    \setlength\tabcolsep{3pt}
    \small
    \center
      \resizebox{0.48\textwidth}{!}{
      \begin{tabular}{llllr}
    \toprule
    \textbf{Model} & \textbf{Initialisation} & \textbf{Visual Tokens} & \textbf{Pretrain Data} &\textbf{\# Params} \\
    \midrule
    \multirow{3}{*}{mUNITER} & \multirow{3}{*}{mBERT} & 36 RoIs from & CC (2.7 M) & \multirow{3}{*}{184.83 M} \\
    & & Faster R-CNN & + Wikipedia (104 Langs) & \\
    & & with ResNet-101 & & \\
    \midrule
    \multirow{3}{*}{xUNITER} & \multirow{3}{*}{XLM-R$_\text{BASE}$} & 36 RoIs from & CC (2.7 M) & \multirow{3}{*}{283.76 M} \\
    & & Faster R-CNN & + Wikipedia (104 Langs) & \\
    & & with ResNet-101 & & \\
    \midrule
    \multirow{3}{*}{UC$^2$} & \multirow{3}{*}{XLM-R$_\text{BASE}$} & 36 RoIs from & CC (3.3 M) & \multirow{3}{*}{281.64 M} \\
    & & Faster R-CNN & + NMT CC-\{\textsc{ces},\german,\french\} & \\
    & & with ResNet-101 & + NMT CC-\{\chinese,\japanese\} & \\
    \midrule
    \multirow{3}{*}{M$^3$P} & \multirow{3}{*}{XLM-R$_\text{BASE}$} & 10-100 RoIs from & CC (3.3 M) & \multirow{3}{*}{376.90 M} \\
    & & Faster R-CNN & + Wikipedia (100 Langs) & \\
    & & with ResNeXt-101 & + Panlex (50 Langs) & \\
    \bottomrule
    \end{tabular}
    }
    \vspace{-0.15cm}
\end{table}

\section{Experimental Details} \label{sec:exp-details}

\textbf{Pretrained Models.}
\cref{tab:models} summarises the key properties of the available pretrained multilingual \vl encoders.
They are all single-stream architectures \cite{bugliarello-etal-2021-multimodal}, which mostly differ in the corpora they used for pretraining.

\rparagraph{Experimental Settings}
We train---both for zero-shot fine-tuning and for few-shot experiments---each encoder end-to-end by learning a task-specific `head' network for each dataset on a single NVIDIA V100 (16GB) GPU card.
For each task, we use the same hyperparameters as in the controlled study of~\citet{bugliarello-etal-2021-multimodal} except for maximum sequence length, which we increase to accommodate the multilingual inputs (see details per task below).
We use the AdamW~\citep{loshchilov2018decoupled} optimiser with parameters $\beta_1$=0.9, $\beta_2$=0.999 and $\epsilon$=1e-6.
The base learning rate depends on each task (see below) but it is always first linearly warmed up for 10\% of the task-specific number of steps, and then linearly decayed until up to 20 epochs of training have been reached.
We also apply a weight decay of $10e-4$ and gradient clipping of $1.0$.
Before training, we extract, for each dataset, 36 image regions using a ResNet-101 backbone~\citep{7780459} for mUNITER, xUNITER and UC$^2$; and 10--100 image regions using a ResNeXt-101 backbone~\citep{Xie_2017_CVPR} for M$^3$P---both trained on Visual Genome~\citep{10.1007/s11263-016-0981-7,8578734}.

\rparagraph{Fine-Tuning Setup}
To increase accessibility to the benchmark by practitioners with limited computing resources, we fix the number of epochs for which we train our baselines so that they can be trained in less than 12 hours per dataset.\footnote{Except for M$^3$P, which is twice as big as the other models.}
In particular, we set up each task based on prior work~\cite{Lu_2020_CVPR,bugliarello-etal-2021-multimodal} as follows.
\begin{itemize}[noitemsep, topsep=1pt]
    \item \textbf{Visual Entailment.} We treat XVNLI as a three-way classification problem, and apply a linear layer from the pooled representation of each Transformer encoder. We fine-tune each model using a binary cross-entropy loss with a batch size of 128 for 10 epochs. We use a learning rate of 2e-5 and max token length of 80.
    \item \textbf{Visual QA.} As in~\citet{Hudson_2019_CVPR}, we treat GQA as a multi-label classification task by assigning a score to each answer according to its relevancy to the ground truth answer. We apply a two-layer MLP with a GeLU activation~\citep{hendrycks2016gelu} function in between on top of the pooled representation. We fine-tune on the \emph{balanced} split of GQA using a binary cross-entropy loss against 1{,}842 labels with a batch size of 256 for 5 epochs. We use a learning rate of 4e-5 and max token length of 40.
    \item \textbf{Visual Reasoning.} We cast this task as a binary classification problem: Given the two pooled representations from encoding the textual description with each of the two images independently, we concatenate and feed them into an MLP (same as for GQA) to predict a True or False label. We fine-tune using a binary cross-entropy loss with a batch size of 64 for 20 epochs. We use a learning rate of 1e-5 and max token length of 80.
\item \textbf{Image--Text Retrieval.} We train a four-way linear classifier against the pooled representations from the true image-caption pair and three hard negatives determined offline. We optimise our models using a cross-entropy loss, except for UC$^2$ for which we found the authors' triplet loss to work better (see \cref{sec:supplementary}). We train with a batch size of 64 for 10 epochs for xFlickr\&CO; and for 2 epochs on the larger WIT. We use a learning rate of 2e-5 and max token length of 80.
\end{itemize}

\rparagraph{Few-shot Setup}
We follow the same setup as in the above described fine-tuning procedure, with the exception of searching across three values for the base learning rate \{1e-5, 5e-5, 1e-4\} and always training for 20 epochs. 
The batch size is adjusted according to the number of data points available for a given split (\cref{tab:few-shot-bs}), and models are initialised from the English fine-tuned parameter set.
For each run, we select the checkpoint that achieves the largest validation performance for evaluation on the test sets.

While the validation sets are in English for every dataset except for xGQA, we found no significant difference in using machine translations of each validation set for checkpoint selection. 
Therefore, we decided to report results using the original validation sets as this more closely matches the few-shot scenario.
Nevertheless, we release the corresponding translations for each dataset for future extensions. 
\begin{table}[t]
\caption{Few-shot batch sizes.}
\label{tab:few-shot-bs}
\vskip 0.10in
\setlength\tabcolsep{3pt}
\small
\center
  \begin{tabular}{llrrrrrr}
\toprule
\textbf{Dataset} & \textbf{Metric} & & & & & & \\
\midrule
\multirow{2}{*}{\small \shortstack{XVNLI}} & \# shots & 1 & 5 & 10 & 20 & 25 & 48 \\
& batch size & 2 & 16 & 32 & 64 & 64 & 64 \\
\midrule
\multirow{2}{*}{\small \shortstack{xGQA}} & \# shots & 1 & 5 & 10 & 20 & 25 & 48 \\
& batch size & 8 & 16 & 32 & 64 & 64 & 64 \\
\midrule
\multirow{2}{*}{\small \shortstack{MaRVL}} & \# shots & 1 & 2 & 4 & 10 & 10$\times$2 & \\
& batch size & 2 & 4 & 8 & 16 & 32 & \\
\midrule
\multirow{2}{*}{\small \shortstack{xFlickr\&CO}} & \# shots & 1 & 5 & 10 & 25 & 50 & 100 \\
& batch size & 1 & 2 & 4 & 8 & 16 & 32 \\
\bottomrule
\end{tabular}
\vskip 0.05in
    \caption{Results of reimplemented models on Multi30K. Mean Recall for zero-shot evaluation (i.e. English-only fine-tuning).}
    \label{tab:multi30k}
    \vskip 0.10in
    \centering
    \small
    \begin{tabular}{l c c c c}
    \toprule
        \textbf{Method} & \textbf{\english} & \textbf{\german} & \textbf{\french} & \textbf{\czech} \\
    \midrule
        M$^3$P~\cite{Ni_2021_CVPR} & 87.4 & 58.5 & 46.0 & 36.8 \\
        UC$^2$~\cite{Zhou_2021_CVPR} & 87.2 & 74.9 & 74.0 & 67.9 \\
    \midrule
        M$^3$P & 83.1 & 55.5 & 51.9 & 45.4 \\
        UC$^2$ & 82.7 & 69.9 & 71.1 & 65.7 \\
        mUNITER & 83.2 & 43.7 & 47.3 & 21.0 \\
        xUNITER & 81.8 & 50.2 & 48.4 & 37.5 \\
    \bottomrule
    \end{tabular}
\vskip 0.05in
    \caption{Results of reimplemented models on xGQA. Accuracy for zero-shot evaluation (i.e. English-only fine-tuning).}
    \label{tab:xgqa-zs}
    \vskip 0.10in
    \centering
    \small
      \resizebox{0.48\textwidth}{!}{
    \begin{tabular}{lccccccccc}
    \toprule
        \textbf{Method} & \textbf{\english} & \textbf{\bengali} & \textbf{\german} & \textbf{\indonesian} & \textbf{\korean} & \textbf{\portuguese} & \textbf{\russian} & \textbf{\chinese} & \textbf{avg} \\
    \midrule
    M$^3$P & 58.4 & 15.8 & 23.9 & 22.6 & 16.9 & 24.4 & 20.4 & 18.6 & 20.4 \\
    OSCAR+$^\text{Emb}$ & 62.2 & 14.9 & 17.3 & 18.3 & 17.1 & 19.2 & 10.5 & 16.4 & 16.3 \\
    OSCAR+$^\text{Ada}$ & 60.3 & 15.4 & 18.9 & 18.8 & 15.3 & 27.0 & 17.5 & 15.0 & 18.3 \\
    mBERT$^\text{Ada}$ & 56.2 & 15.1 & 29.8 & 19.2 & 19.1 & 30.4 & 24.4 & 24.9 & 23.2 \\
    \midrule
    M$^3$P & 57.9 & 19.2 & 35.6 & 29.6 & 28.5 & 35.7 & 32.9 & 31.1 & 33.8 \\
    UC$^2$ & 59.4 & 20.8 & 45.1 & 28.1 & 24.3 & 27.6 & 31.9 & 33.9 & 33.9 \\
    mUNITER & 58.2 & 3.9 & 26.3 & 13.0 & 5.5 & 17.4 & 9.3 & 8.4 & 17.7 \\
    xUNITER & 57.7 & 9.7 & 35.7 & 36.3 & 14.6 & 24.7 & 20.8 & 19.2 & 27.3 \\
    \bottomrule
    \end{tabular}
    }
\end{table}

\vspace{-1mm}
\section{Reproducibility and Additional Results} \label{sec:supplementary}

We provide more details regarding our experimental setup in order to ensure reproducibility (\S\ref{sec:repr_results}). Further, we report additional per-language (\S\ref{sec:perlanguage}) and few-shot (\S\ref{sec:more-few-results}) results, which supplement the ones provided in the main paper.

\vspace{-1mm}
\subsection{A Unified Implementation Framework}\label{sec:repr_results}
In order to provide a more lightweight setup for all the experiments in \benchmark, which concerns the experiments we conducted for this work as well as any future experiments by other researchers, we reimplemented M$^3$P and UC$^2$ in a unified framework based on \volta~\citep{bugliarello-etal-2021-multimodal}, which already implements mUNITER and xUNITER, along with five English pretrained encoders. In order to verify the correctness of the implementation, we evaluated each model on Multi30K~\citep{elliott-etal-2016-multi30k} in a zero-shot scenario.
The results shown in~\cref{tab:multi30k} prove the correctness of our reimplementation: Performance of both models is within a couple of points from the one reported by the original authors.
In particular, we obtain similar results as~\citet{Geigle:2022tacl} for M$^3$P. 
Moreover, we found that using a cross-entropy loss for UC$^2$ led to a drop in performance of approximately 2.5 mean Recall points in each language. Similarly, using a triplet loss for M$^3$P led to a drop in performance of nearly 3 points.

In~\cref{tab:xgqa-zs}, we report the performance of our models on the zero-shot splits of xGQA, for a direct comparison with the standard GQA test split and the performance achieved by~\citet{pfeiffer2021xgqa}.
We find that our setup leads to significantly better performance: our best model, UC$^2$ gains over 10 points compared to the best model from~\citet{pfeiffer2021xgqa}.
In addition, our M$^3$P model is 13.4\% more accurate than the one trained by the authors. 
We believe the main reason for this considerable performance gain is our deeper prediction head: While the authors place a linear classification head, our models use a non-linear two-layer feed-forward head on top of the Transformer representations. 
\begin{table}[!t]
\vskip 0.05in
\caption{Zero-shot per-language results on WIT (IR and TR).}
\label{tab:wit}
\vskip 0.10in
\setlength\tabcolsep{3pt}
\small
\center
  \resizebox{0.48\textwidth}{!}{
  \begin{tabular}{llrrrrrrrrrr}
\toprule
\multicolumn{2}{l}{\textbf{WIT}} & \multicolumn{10}{c}{\textbf{Language}} \\
\textbf{Type} & \textbf{Model} & \multicolumn{1}{c}{\textbf{\arab}} & \multicolumn{1}{c}{\textbf{\bulgarian}} & \multicolumn{1}{c}{\textbf{\danish}} & \multicolumn{1}{c}{\textbf{\greek}} & \multicolumn{1}{c}{\textbf{\estonian}} & \multicolumn{1}{c}{\textbf{\indonesian}} & \multicolumn{1}{c}{\textbf{\japanese}} & \multicolumn{1}{c}{\textbf{\korean}} & \multicolumn{1}{c}{\textbf{\turkish}} & \multicolumn{1}{c}{\textbf{\vietnamese}} \\
\midrule
\multirow{4}{*}{\small \shortstack{IR}} & mUNITER & 7.74 & 8.26 & \textbf{10.66} & 8.95 & \textbf{7.67} & \textbf{10.88} & 9.00 & 5.91 & \textbf{9.57} & \textbf{13.00} \\
& xUNITER & 7.63 & 8.49 & 10.32 & \textbf{11.23} & 6.41 & 10.21 & 7.30 & \textbf{6.34} & \textbf{9.57} & 9.72 \\
& UC$^2$ & 6.62 & \textbf{8.84} & 9.43 & 8.77 & 4.69 & 9.88 & \textbf{9.80} & 4.30 & 7.49 & 8.46 \\
& M$^3$P & \textbf{8.87} & \textbf{8.84} & 9.43 & 9.65 & 5.38 & 8.66 & 7.00 & 6.12 & 6.52 & 10.78 \\
\midrule
\multirow{4}{*}{\small \shortstack{TR}} & mUNITER & \textbf{9.21} & 10.17 & \textbf{12.16} & 10.54 & \textbf{8.33} & \textbf{12.88} & 8.79 & 6.75 & \textbf{10.87} & \textbf{15.07} \\
& xUNITER & 9.08 & \textbf{10.30} & 9.34 & \textbf{12.38} & 7.82 & 10.66 & 10.10 & 6.97 & 9.69 & 11.74 \\
& UC$^2$ & 8.32 & 7.69 & 10.44 & 11.64 & 6.03 & 11.47 & \textbf{10.81} & 5.74 & 8.81 & 9.90 \\
& M$^3$P & 8.32 & 9.80 & 11.79 & 12.02 & 8.21 & 10.89 & 8.43 & \textbf{7.09} & 10.57 & 12.66 \\
\bottomrule
\end{tabular}
}
\vskip 0.05in
\caption{Full per-language results on XVNLI.}
\label{tab:xvnli}
\vskip 0.10in
\setlength\tabcolsep{3pt}
\small
\center
  \resizebox{0.48\textwidth}{!}{
  \begin{tabular}{llccccccc}
\toprule
\multicolumn{2}{l}{\textbf{XVNLI}} & \multicolumn{7}{c}{\textbf{\# shots}} \\
\textbf{Lang} & \textbf{Model} & \textbf{0} & \textbf{1} & \textbf{5} & \textbf{10} & \textbf{20} & \textbf{25} & \textbf{48} \\
\midrule
\multirow{4}{*}{\small \shortstack{\arab}} & mUNITER & 46.73 & 46.99 & 46.39 & 49.40 & 47.16 & 48.97 & 46.91 \\
& xUNITER & 51.98 & 52.32 & 52.32 & 54.81 & 54.55 & 53.95 & 54.04 \\
& UC$^2$ & \textbf{56.19} & \textbf{56.36} & \textbf{57.65} & \textbf{57.82} & \textbf{59.11} & \textbf{58.51} & \textbf{56.87} \\
& M$^3$P & 55.24 & 54.98 & 55.07 & 56.19 & 55.24 & 56.79 & 56.01 \\
\midrule
\multirow{4}{*}{\small \shortstack{\spanish}} & mUNITER & 56.96 & 56.79 & 57.47 & 57.30 & 57.73 & 57.65 & 57.73 \\
& xUNITER & \textbf{58.94} & \textbf{59.02} & {58.94} & \textbf{60.74} & 59.88 & 59.11 & 60.22 \\
& UC$^2$ & 57.47 & 57.65 & \textbf{61.34} & 59.79 & \textbf{62.63} & \textbf{59.79} & \textbf{62.80} \\
& M$^3$P & 58.85 & 58.33 & {58.94} & 60.05 & 59.19 & 59.28 & 60.40 \\
\midrule
\multirow{4}{*}{\small \shortstack{\french}} & mUNITER & 59.36 & 59.45 & 59.19 & 59.54 & 59.36 & 59.62 & 59.36 \\
& xUNITER & 63.32 & 63.49 & 64.26 & 63.83 & 64.26 & 64.09 & 64.52 \\
& UC$^2$ & \textbf{69.67} & \textbf{69.67} & \textbf{69.67} & \textbf{69.76} & \textbf{69.84} & \textbf{69.76} & \textbf{69.76} \\
& M$^3$P & 56.36 & 56.62 & 57.99 & 58.33 & 57.82 & 57.65 & 58.59 \\
\midrule
\multirow{4}{*}{\small \shortstack{\russian}} & mUNITER & 51.72 & 51.72 & 51.46 & 50.52 & 52.32 & 51.46 & 51.80 \\
& xUNITER & 59.71 & 59.54 & 59.54 & 59.54 & 61.94 & 61.34 & 63.40 \\
& UC$^2$ & \textbf{64.86} & \textbf{64.43} & \textbf{64.52} & \textbf{64.95} & \textbf{65.38} & \textbf{65.38} & \textbf{65.29} \\
& M$^3$P & 62.54 & 62.46 & 62.72 & 63.23 & 62.46 & 62.97 & 62.46 \\
\bottomrule
\end{tabular}
}
\end{table}

\begin{table}[!t]
\caption{Full per-language results on xGQA.}
\label{tab:xgqa}
\setlength\tabcolsep{3pt}
\small
\center
  \resizebox{0.48\textwidth}{!}{
  \begin{tabular}{llccccccc}
\toprule
\multicolumn{2}{l}{\textbf{xGQA}} & \multicolumn{7}{c}{\textbf{\# shots}} \\
\textbf{Lang} & \textbf{Model} & \textbf{0} & \textbf{1} & \textbf{5} & \textbf{10} & \textbf{20} & \textbf{25} & \textbf{48} \\
\midrule
\multirow{4}{*}{\small \shortstack{\bengali}} & mUNITER & 3.06 & 19.36 & 23.94 & 27.53 & 30.04 & 31.04 & 34.89 \\
& xUNITER & 10.80 & \textbf{23.92} & 29.43 & 31.67 & 34.84 & \textbf{36.18} & 37.55 \\
& UC$^2$ & \textbf{19.99} & 22.52 & 30.96 & 32.84 & 35.69 & 35.12 & \textbf{38.90} \\
& M$^3$P & 18.64 & 23.42 & \textbf{31.07} & \textbf{33.37} & \textbf{35.74} & {35.94} & 37.76 \\
\midrule
\multirow{4}{*}{\small \shortstack{\german}} & mUNITER & 23.95 & 29.43 & 33.88 & 35.40 & 37.82 & 37.43 & 40.29 \\
& xUNITER & 34.83 & 38.44 & 39.71 & 40.97 & 41.93 & 42.19 & 43.60 \\
& UC$^2$ & \textbf{42.85} & \textbf{43.76} & \textbf{44.68} & \textbf{45.72} & \textbf{46.70} & \textbf{47.24} & \textbf{48.18} \\
& M$^3$P & 33.42 & 34.37 & 39.66 & 40.73 & 41.78 & 41.93 & 43.19 \\
\midrule
\multirow{4}{*}{\small \shortstack{\indonesian}} & mUNITER & 9.36 & 27.31 & 31.39 & 33.28 & 36.49 & 35.85 & 38.06 \\
& xUNITER & \textbf{33.73} & \textbf{35.28} & 37.38 & 37.96 & 39.48 & 39.40 & 40.90 \\
& UC$^2$ & 28.67 & 34.76 & \textbf{38.95} & \textbf{40.11} & \textbf{41.26} & 41.51 & \textbf{43.11} \\
& M$^3$P & 32.48 & 33.11 & 38.14 & 39.84 & 40.51 & \textbf{41.53} & 41.38 \\
\midrule
\multirow{4}{*}{\small \shortstack{\korean}} & mUNITER & 4.21 & 19.40 & 26.47 & 29.03 & 31.95 & 32.76 & 35.23 \\
& xUNITER & 12.12 & 23.45 & 31.49 & 34.70 & 36.66 & 37.26 & 39.32 \\
& UC$^2$ & 21.36 & 29.33 & 33.02 & 34.50 & 36.09 & \textbf{38.40} & \textbf{39.63} \\
& M$^3$P & \textbf{25.11} & \textbf{29.74} & \textbf{34.53} & \textbf{35.77} & \textbf{37.11} & 37.81 & 38.58 \\
\midrule
\multirow{4}{*}{\small \shortstack{\portuguese}} & mUNITER & 13.67 & 22.88 & 31.09 & 34.38 & 37.05 & 37.32 & 39.41 \\
& xUNITER & 22.13 & 30.10 & 36.44 & 38.72 & 39.73 & 41.06 & 42.56 \\
& UC$^2$ & 30.42 & 32.10 & \textbf{39.42} & \textbf{39.57} & \textbf{41.73} & 41.27 & \textbf{43.23} \\
& M$^3$P & \textbf{31.40} & \textbf{33.37} & 37.62 & 39.47 & 41.09 & \textbf{41.96} & 43.01 \\
\midrule
\multirow{4}{*}{\small \shortstack{\russian}} & mUNITER & 8.49 & 22.51 & 29.42 & 31.98 & 32.90 & 34.34 & 34.78 \\
& xUNITER & 18.84 & 27.32 & 34.05 & 36.39 & 38.19 & 38.37 & 39.40 \\
& UC$^2$ & \textbf{31.00} & 33.04 & \textbf{37.40} & \textbf{38.13} & \textbf{40.63} & \textbf{41.09} & \textbf{42.79} \\
& M$^3$P & 27.50 & \textbf{33.50} & 37.23 & 37.86 & 39.18 & 39.76 & 42.15 \\
\midrule
\multirow{4}{*}{\small \shortstack{\chinese}} & mUNITER & 7.03 & 18.33 & 30.90 & 32.25 & 36.18 & 35.52 & 37.81 \\
& xUNITER & 19.55 & 31.36 & 37.35 & 37.34 & 39.19 & 39.48 & 41.42 \\
& UC$^2$ & \textbf{31.16} & \textbf{37.54} & \textbf{41.21} & \textbf{41.49} & \textbf{43.46} & \textbf{43.58} & \textbf{44.82} \\
& M$^3$P & 28.65 & 30.67 & 36.77 & 37.77 & 39.25 & 39.99 & 41.19 \\
\bottomrule
\end{tabular}
}
\vskip 0.05in
\caption{Full per-language results on MaRVL.}
\label{tab:marvl}
\setlength\tabcolsep{3pt}
\small
\center
  \resizebox{0.48\textwidth}{!}{
  \begin{tabular}{llcccccc}
\toprule
\multicolumn{2}{l}{\textbf{MaRVL}} & \multicolumn{6}{c}{\textbf{\# shots}} \\
\textbf{Lang} & \textbf{Model} & \textbf{0} & \textbf{1} & \textbf{2} & \textbf{4} & \textbf{10} & \textbf{10$\times$2} \\
\midrule
\multirow{4}{*}{\small \shortstack{\indonesian}} & mUNITER & 54.79 & 51.42 & 52.13 & 54.34 & 56.83 & 51.42 \\
& xUNITER & 55.14 & \textbf{56.12} & \textbf{57.18} & \textbf{58.87} & \textbf{58.87} & \textbf{58.60} \\
& UC$^2$ & \textbf{56.74} & \textbf{56.12} & 56.29 & 57.53 & 57.62 & 58.51 \\
& M$^3$P & 56.47 & 49.02 & 48.05 & 50.18 & 50.09 & 50.09 \\
\midrule
\multirow{4}{*}{\small \shortstack{\swahili}} & mUNITER & 51.17 & - & - & - & - & - \\
& xUNITER & 55.51 & - & - & - & - & - \\
& UC$^2$ & 52.62 & - & - & - & - & - \\
& M$^3$P & \textbf{55.69} & - & - & - & - & - \\
\midrule
\multirow{4}{*}{\small \shortstack{\tamil}} & mUNITER & 52.66 & - & - & - & - & - \\
& xUNITER & 53.06 & - & - & - & - & - \\
& UC$^2$ & \textbf{60.47} & - & - & - & - & - \\
& M$^3$P & 56.04 & - & - & - & - & - \\
\midrule
\multirow{4}{*}{\small \shortstack{\turkish}} & mUNITER & 54.66 & 52.54 & 55.93 & 55.59 & 54.66 & 54.75 \\
& xUNITER & 56.19 & \textbf{55.93} & \textbf{57.46} & \textbf{57.54} & \textbf{57.80} & \textbf{58.05} \\
& UC$^2$ & 56.70 & 55.76 & 54.49 & 52.63 & 55.93 & 56.27 \\
& M$^3$P & \textbf{56.78} & 51.19 & 49.32 & 49.58 & 49.41 & 49.58 \\
\midrule
\multirow{4}{*}{\small \shortstack{\chinese}} & mUNITER & 55.34 & 56.52 & 54.84 & 54.55 & 55.34 & 54.05 \\
& xUNITER & 53.06 & 54.45 & 53.66 & 54.94 & 53.46 & 55.73 \\
& UC$^2$ & \textbf{59.88} & \textbf{58.99} & \textbf{57.02} & \textbf{58.99} & \textbf{57.21} & \textbf{60.18} \\
& M$^3$P & 55.04 & 50.69 & 50.40 & 50.20 & 49.80 & 49.70 \\
\bottomrule
\end{tabular}
}
\end{table}

\begin{table}[t]
\caption{Full per-language results on xFlickr\&CO IR.}
\label{tab:xflickrco-ir}
\vskip 0.05in
\setlength\tabcolsep{3pt}
\small
\center
  \resizebox{0.48\textwidth}{!}{
  \begin{tabular}{llrrrrrrr}
\toprule
\multicolumn{2}{l}{\textbf{xFlickr\&CO IR}} & \multicolumn{7}{c}{\textbf{\# shots}} \\
\textbf{Lang} & \textbf{Model} & \multicolumn{1}{c}{\textbf{0}} & \multicolumn{1}{c}{\textbf{1}} & \multicolumn{1}{c}{\textbf{5}} & \multicolumn{1}{c}{\textbf{10}} & \multicolumn{1}{c}{\textbf{25}} & \multicolumn{1}{c}{\textbf{50}} & \multicolumn{1}{c}{\textbf{100}} \\
\midrule
\multirow{4}{*}{\small \shortstack{\german}} & mUNITER & 12.05 & 10.50 & 11.90 & 11.95 & 11.95 & 11.95 & 12.05 \\
& xUNITER & 14.55 & 14.40 & 14.35 & 14.45 & 14.45 & 14.45 & 14.55 \\
& UC$^2$ & \textbf{28.60} & \textbf{27.40} & \textbf{28.15} & \textbf{28.15} & \textbf{28.15} & \textbf{28.15} & \textbf{28.60} \\
& M$^3$P & 13.35 & 12.25 & 12.85 & 11.95 & 13.15 & 13.35 & 13.25 \\
\midrule
\multirow{4}{*}{\small \shortstack{\spanish}} & mUNITER & 13.15 & 13.70 & 14.25 & 14.25 & 14.25 & 14.25 & 13.15 \\
& xUNITER & \textbf{16.10} & \textbf{16.10} & \textbf{16.05} & \textbf{16.05} & \textbf{16.05} & \textbf{16.05} & \textbf{16.10} \\
& UC$^2$ & 15.95 & 15.40 & 15.70 & 15.50 & 8.55 & 15.40 & 15.95 \\
& M$^3$P & 13.40 & 13.40 & 13.80 & 13.95 & 13.50 & 13.40 & 13.40 \\
\midrule
\multirow{4}{*}{\small \shortstack{\indonesian}} & mUNITER & 5.95 & 6.05 & 4.55 & 5.75 & 5.65 & 5.80 & 5.95 \\
& xUNITER & \textbf{16.50} & \textbf{16.90} & \textbf{16.55} & \textbf{16.20} & \textbf{16.60} & \textbf{16.60} & \textbf{16.50} \\
& UC$^2$ & 14.60 & 14.50 & 14.55 & 12.30 & 14.70 & 14.65 & 14.60 \\
& M$^3$P & 13.20 & 13.20 & 14.20 & 14.15 & 13.50 & 13.20 & 13.20 \\
\midrule
\multirow{4}{*}{\small \shortstack{\japanese}} & mUNITER & 6.30 & 6.20 & 6.20 & 6.20 & 6.20 & 6.20 & 6.30 \\
& xUNITER & 10.25 & 11.50 & 11.80 & 11.95 & 10.65 & 10.80 & 10.70 \\
& UC$^2$ & \textbf{24.25} & \textbf{23.15} & \textbf{23.00} & \textbf{24.80} & \textbf{24.40} & \textbf{24.35} & \textbf{19.80} \\
& M$^3$P & 10.30 & 11.50 & 10.85 & 12.75 & 11.70 & 12.05 & 11.75 \\
\midrule
\multirow{4}{*}{\small \shortstack{\russian}} & mUNITER & 5.85 & 3.90 & 5.75 & 6.55 & 6.25 & 8.50 & 9.25 \\
& xUNITER & 15.90 & 15.75 & 15.20 & 16.00 & 16.05 & 16.20 & 17.25 \\
& UC$^2$ & \textbf{20.00} & \textbf{20.05} & \textbf{19.90} & \textbf{19.65} & \textbf{21.30} & \textbf{17.65} & \textbf{20.85} \\
& M$^3$P & 15.95 & 15.95 & 17.10 & 16.15 & 16.75 & 16.40 & 16.65 \\
\midrule
\multirow{4}{*}{\small \shortstack{\turkish}} & mUNITER & 1.75 & 1.90 & 1.40 & 1.30 & 1.30 & 1.35 & 1.75 \\
& xUNITER & \textbf{9.05} & \textbf{9.35} & \textbf{9.05} & \textbf{9.05} & \textbf{9.05} & \textbf{9.05} & \textbf{9.05} \\
& UC$^2$ & 7.15 & 7.20 & 7.05 & 5.55 & 6.85 & 4.90 & 7.15 \\
& M$^3$P & 7.75 & 7.75 & 7.45 & 7.45 & 7.35 & 7.75 & 7.75 \\
\midrule
\multirow{4}{*}{\small \shortstack{\chinese}} & mUNITER & 11.35 & 11.50 & 11.45 & 11.40 & 11.45 & 11.45 & 11.35 \\
& xUNITER & 15.95 & 17.50 & 7.90 & 17.60 & 16.00 & 13.65 & 15.95 \\
& UC$^2$ & \textbf{31.60} & \textbf{31.30} & \textbf{31.05} & \textbf{29.75} & \textbf{31.40} & \textbf{25.50} & \textbf{31.60} \\
& M$^3$P & 16.45 & 14.85 & 16.15 & 9.30 & 16.40 & 12.65 & 16.45 \\
\bottomrule
\end{tabular}
}
\end{table}

\begin{table}[t]
\caption{Full per-language results on xFlickr\&CO TR.}
\label{tab:xflickrco-tr}
\vskip 0.05in
\setlength\tabcolsep{3pt}
\small
\center
  \resizebox{0.48\textwidth}{!}{
  \begin{tabular}{llrrrrrrr}
\toprule
\multicolumn{2}{l}{\textbf{xFlickr\&CO TR}} & \multicolumn{7}{c}{\textbf{\# shots}} \\
\textbf{Lang} & \textbf{Model} & \multicolumn{1}{c}{\textbf{0}} & \multicolumn{1}{c}{\textbf{1}} & \multicolumn{1}{c}{\textbf{5}} & \multicolumn{1}{c}{\textbf{10}} & \multicolumn{1}{c}{\textbf{25}} & \multicolumn{1}{c}{\textbf{50}} & \multicolumn{1}{c}{\textbf{100}} \\
\midrule
\multirow{4}{*}{\small \shortstack{\german}} & mUNITER & 11.85 & 10.65 & 11.70 & 11.70 & 11.70 & 11.65 & 11.85 \\
& xUNITER & 13.25 & 12.95 & 13.20 & 13.15 & 13.15 & 13.15 & 13.25 \\
& UC$^2$ & \textbf{23.90} & \textbf{22.25} & \textbf{23.45} & \textbf{23.45} & \textbf{23.45} & \textbf{23.45} & \textbf{23.90} \\
& M$^3$P & 11.85 & 10.65 & 12.35 & 10.60 & 12.60 & 11.85 & 12.55 \\
\midrule
\multirow{4}{*}{\small \shortstack{\spanish}} & mUNITER & 13.05 & 14.70 & 14.05 & 14.05 & 14.05 & 14.05 & 13.05 \\
& xUNITER & 15.10 & 15.60 & 15.30 & 15.30 & \textbf{15.30} & 15.30 & 15.10 \\
& UC$^2$ & \textbf{15.30} & \textbf{15.85} & \textbf{16.15} & \textbf{16.15} & 9.25 & \textbf{15.45} & \textbf{15.30} \\
& M$^3$P & 12.15 & 12.15 & 11.60 & 11.60 & 11.55 & 12.15 & 12.15 \\
\midrule
\multirow{4}{*}{\small \shortstack{\indonesian}} & mUNITER & 7.55 & 8.30 & 4.85 & 6.30 & 6.35 & 6.20 & 7.55 \\
& xUNITER & \textbf{16.75} & \textbf{17.05} & \textbf{16.85} & \textbf{16.65} & \textbf{16.75} & \textbf{16.80} & \textbf{16.75} \\
& UC$^2$ & 13.60 & 14.90 & 14.20 & 13.70 & 14.05 & 13.95 & 13.60 \\
& M$^3$P & 12.10 & 12.10 & 13.00 & 13.15 & 12.20 & 12.10 & 12.10 \\
\midrule
\multirow{4}{*}{\small \shortstack{\japanese}} & mUNITER & 7.70 & 8.10 & 7.55 & 7.55 & 7.55 & 7.55 & 7.70 \\
& xUNITER & 9.85 & 11.80 & 11.30 & 11.45 & 10.10 & 10.35 & 10.15 \\
& UC$^2$ & \textbf{22.40} & \textbf{21.20} & \textbf{20.25} & \textbf{22.20} & \textbf{22.45} & \textbf{22.25} & \textbf{18.40} \\
& M$^3$P & 9.65 & 12.30 & 10.60 & 11.50 & 11.35 & 11.25 & 10.50 \\
\midrule
\multirow{4}{*}{\small \shortstack{\russian}} & mUNITER & 6.80 & 7.60 & 7.50 & 7.15 & 7.25 & 8.55 & 10.00 \\
& xUNITER & 14.80 & 14.65 & 15.10 & 14.50 & 15.25 & 14.60 & 14.70 \\
& UC$^2$ & \textbf{16.75} & \textbf{17.00} & \textbf{16.75} & \textbf{17.75} & \textbf{17.50} & \textbf{16.30} & \textbf{18.65} \\
& M$^3$P & 14.45 & 14.45 & 14.65 & 14.90 & 14.40 & 15.85 & 15.40 \\
\midrule
\multirow{4}{*}{\small \shortstack{\turkish}} & mUNITER & 3.25 & 3.80 & 2.45 & 2.10 & 1.95 & 2.00 & 3.25 \\
& xUNITER & \textbf{10.05} & \textbf{10.30} & \textbf{9.95} & \textbf{9.95} & \textbf{9.95} & \textbf{9.95} & \textbf{10.05} \\
& UC$^2$ & 6.95 & 7.15 & 7.00 & 6.60 & 7.10 & 6.70 & 6.95 \\
& M$^3$P & 8.35 & 8.35 & 8.10 & 8.10 & 8.05 & 8.35 & 8.35 \\
\midrule
\multirow{4}{*}{\small \shortstack{\chinese}} & mUNITER & 11.85 & 12.40 & 11.65 & 11.70 & 11.65 & 11.65 & 11.85 \\
& xUNITER & 14.80 & 15.95 & 8.55 & 15.95 & 14.95 & 13.40 & 14.80 \\
& UC$^2$ & \textbf{26.30} & \textbf{25.80} & \textbf{25.75} & \textbf{24.40} & \textbf{25.20} & \textbf{21.50} & \textbf{26.30} \\
& M$^3$P & 14.75 & 13.65 & 14.40 & 10.65 & 15.00 & 12.90 & 14.75 \\
\bottomrule
\end{tabular}
}
 	\centering
 	\includegraphics[width=0.48\textwidth, trim={3cm 0cm 4cm 0cm}, clip]{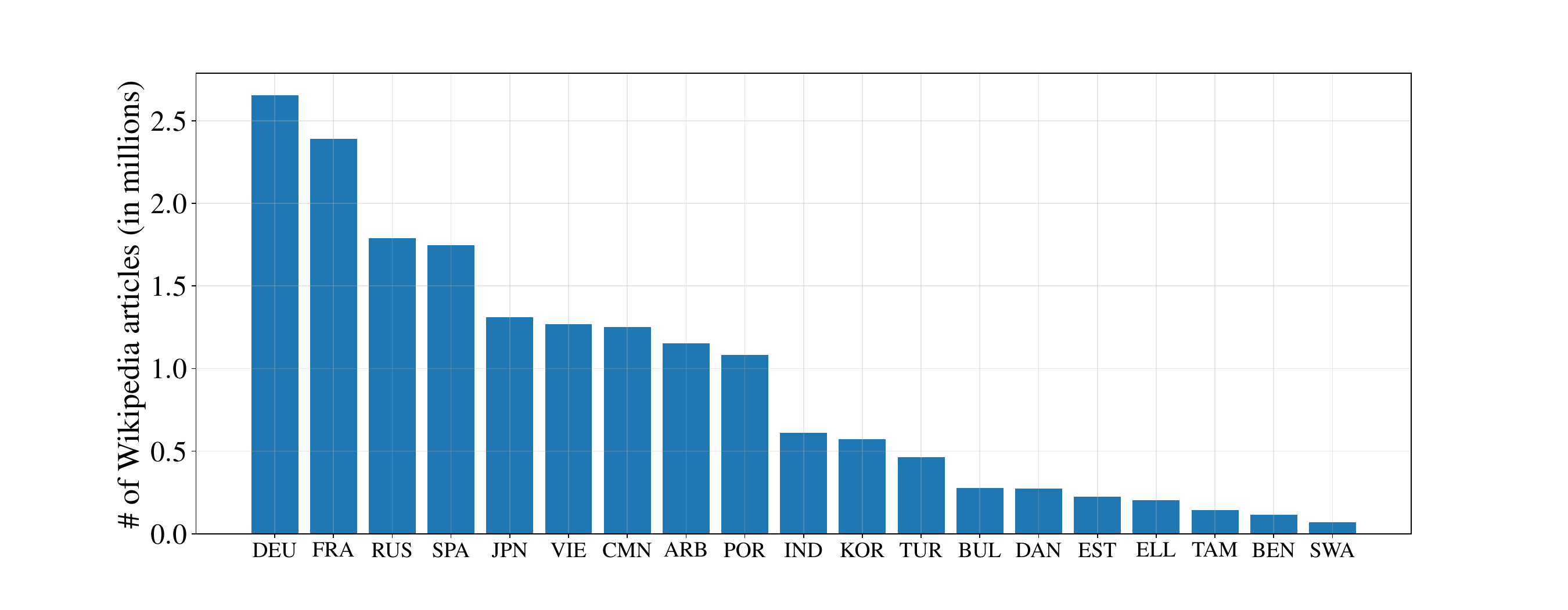}
 	\vspace{-0.9cm}
 	\captionof{figure}{Number of Wikipedia articles per \benchmark{} language.} 
 	\label{fig:wiki-sizes}
\end{table}

\subsection{Performance per Target Language}\label{sec:perlanguage}
\cref{tab:xvnli,tab:xgqa,tab:marvl,tab:xflickrco-ir,tab:xflickrco-tr,tab:wit} present language-specific results for each dataset and number of shots.
Retrieval performance in WIT is relatively constant across languages, likely due to the different distribution of captions and images in the Wikipedia domain.
Different from other datasets (see below), this is true even for UC$^2$ in the \japanese{} split, even though the model was pretrained on large-scale multimodal data in Japanese.

In XVNLI and xGQA, performance peaks for French and German, respectively.
While typologically more similar and geographically closer to English, we find that the amount of multilingual data, shown in \cref{fig:wiki-sizes}, is also indicative of the final performance.
For instance, both mUNITER and xUNITER---which were pretrained on the same data but with different multilingual sampling ratios during text-only and multimodal pretraining---perform on par or better in Russian than in Spanish and Portuguese, respectively.
Moreover, UC$^2$ consistently achieves the highest performance in French, German and Mandarin, which are 3/5 Conceptual Captions translations on which the model was pretrained.

On MaRVL, this is also the case for the Mandarin split, although UC$^2$ now reaches lower gains relative to other models in the other datasets.
This shows how the out-of-distribution nature of the images in MaRVL limits the efficacy of `translate train' approaches for zero-shot transfer here.
Interestingly, Mandarin is also the language in which the other models perform the worst, indicating that large-scale multimodal data in the target language can still benefit downstream performance.
Overall, due to the domain shifts present in MaRVL, the performance across all languages and models is comparably low.

Although being in-distribution, performance in xFlickr\&CO is consistently low in every language, and especially in Turkish.
Once more, UC$^2$ largely outperforms the other systems in \german, \japanese{} and \chinese{} but is on par in the other languages.

All in all, the resulting differences between the best- and worst-performing languages are considerable, amounting 13.5 points for XVNLI, 22.9 points for xGQA, and 22.5 points for xFlickr\&Co, even though the visual input is language-independent in these evaluation sets.

\subsection{Additional Results for Few-shot Learning}\label{sec:more-few-results}

\begin{table}[t]
\caption{Few-shot AUC (in $\%$) averaged across languages.} \label{tab:few-shot-auc}
\vskip 0.10in
\setlength\tabcolsep{2pt}
\small
\center
  \begin{tabular}{lrrcrr}
\toprule
\multirow{2}{*}{\small \shortstack{\textbf{Model}}}         &  \multicolumn{1}{c}{\textbf{NLI}}   & \multicolumn{1}{c}{\textbf{QA}} & \multicolumn{1}{c}{\textbf{Reasoning}}  & \multicolumn{2}{c}{\textbf{Retrieval}} \\
  \cmidrule(r){2-2}
  \cmidrule(r){3-3}
  \cmidrule(r){4-4}
  \cmidrule(r){5-6}
 & \multicolumn{1}{c}{XVNLI} & \multicolumn{1}{c}{xGQA}     & \multicolumn{1}{c}{MaRVL} & \multicolumn{2}{c}{xFlickr\&CO}    \\
 & & & & \multicolumn{1}{c}{IR} & \multicolumn{1}{c}{TR} \\
\midrule
mUNITER & 54.10 & 33.67 & 54.68 & 8.36 & 8.89 \\
xUNITER & 59.78 & 38.10 & 56.82 & 14.05 & 13.51 \\
UC$^2$ & \textbf{63.43} & \textbf{40.43} & \textbf{57.11} & \textbf{19.24} & \textbf{17.31} \\
M$^3$P & 59.07 & 38.89 & 49.96 & 12.90 & 12.07 \\
\bottomrule
\end{tabular}
\end{table}

\rparagraph{Area under the Curve}
To compare the overall performance of different models, we compute a normalised Area Under Curve (AUC) score for the few-shot setup (plotted in \cref{fig:fewshot-curves-avgV2}).
For a given model $f$ and dataset $D$ with $N_D$ possible shots, each with $s_k$ examples, we approximate AUC according to the trapezoidal rule:
\begin{equation}
    \text{AUC ($f$, $D$)} = \frac{1}{s_{N_D}}\sum_{k=1}^{N_D}{\frac {f(s_{k-1})+f(s_{k})}{2}}\Delta s_{k}.
\end{equation}
The results are listed in \cref{tab:few-shot-auc}. 
In summary, UC$^2$ is by far the strongest model across all tasks; xUNITER comes second and it outperforms M$^3$P (which ranks the third) on reasoning and retrieval; mUNITER performs the worst in general, except for being better than M$^3$P on reasoning.

\begin{table}[t]
\caption{Difference in accuracy between the most-shots and zero-shot setups averaged across languages. For MaRVL, accuracy is averaged across the 3/5 languages with few-shot training data.}
\label{tab:few-shot-diffs}
\vskip 0.10in
\setlength\tabcolsep{4pt}
\small
\center
  \resizebox{0.48\textwidth}{!}{
  \begin{tabular}{lccccc}
\toprule
\multirow{2}{*}{\small \shortstack{\textbf{Model}}}         &  \multicolumn{1}{c}{\textbf{NLI}}   & \multicolumn{1}{c}{\textbf{QA}} & \multicolumn{1}{c}{\textbf{Reasoning}}  & \multicolumn{2}{c}{\textbf{Retrieval}} \\
  \cmidrule(r){2-2}
  \cmidrule(r){3-3}
  \cmidrule(r){4-4}
  \cmidrule(r){5-6}
 & \multicolumn{1}{c}{XVNLI} & \multicolumn{1}{c}{xGQA}     & \multicolumn{1}{c}{MaRVL} & \multicolumn{2}{c}{xFlickr\&CO}    \\
 & & & & \multicolumn{1}{c}{IR} & \multicolumn{1}{c}{TR} \\
\midrule
mUNITER & $\uparrow$ 0.26 & $\uparrow$ 27.24 & $\downarrow$ 0.31 & $\uparrow$ 0.48 & $\uparrow$ 0.46 \\
xUNITER & $\uparrow$ 2.07 & $\uparrow$ 18.96 & $\uparrow$ 2.87 & $\uparrow$ 0.26 & $\uparrow$ 0.03 \\
UC$^2$ & $\uparrow$ 1.63 & $\uparrow$ 13.60 & $\uparrow$ 1.04 & $\downarrow$ 0.52 & $\downarrow$ 0.30 \\
M$^3$P & $\uparrow$ 1.11 & $\uparrow$ 12.87 & $\downarrow$ 6.21 & $\downarrow$ 0.30 & $\uparrow$ 0.36 \\
\bottomrule
\end{tabular}
}
\end{table}

\begin{table}[t]
\caption{Caption density in xFlickr\&CO.}
\label{tab:FlickrCO-num-ir}
\vskip 0.15in
\setlength\tabcolsep{3pt}
\small
\center
  \resizebox{0.48\textwidth}{!}{
  \begin{tabular}{rrr|rr|rr|rr|rr}
\toprule
\multicolumn{3}{l|}{\textbf{xFlickr\&CO}} & \multicolumn{2}{c|}{\textbf{\german{} IR}} & \multicolumn{2}{c|}{\textbf{\japanese{} IR}} & \multicolumn{2}{c|}{\textbf{\german{} TR}} & \multicolumn{2}{c}{\textbf{\japanese{} TR}} \\
\textbf{\# shots} & \textbf{\# img} & \textbf{\# cap/img} & \multicolumn{1}{c}{\textbf{xUNI}} & \multicolumn{1}{c|}{\textbf{UC$^2$}} & \multicolumn{1}{c}{\textbf{xUNI}} & \multicolumn{1}{c|}{\textbf{UC$^2$}} & \multicolumn{1}{c}{\textbf{xUNI}} & \multicolumn{1}{c|}{\textbf{UC$^2$}} & \multicolumn{1}{c}{\textbf{xUNI}} & \multicolumn{1}{c}{\textbf{UC$^2$}} \\
\midrule
\multirow{2}{*}{\small \shortstack{500}} & 500 & 1 & \textbf{17.15} & 27.20 & 11.95 & \textbf{29.35} & \textbf{14.80} & 21.40 & \textbf{13.05} & \textbf{25.45} \\
& 100 & 5 & 13.70 & \textbf{30.60} & \textbf{14.00} & 23.60 & 9.80 & \textbf{25.10} & 12.50 & 21.70 \\
\midrule
\multirow{2}{*}{\small \shortstack{1000}} & 1000 & 1 & 15.50 & \textbf{30.55} & 12.70 & \textbf{28.90} & 14.20 & \textbf{24.65} & \textbf{12.65} & 26.00 \\
& 200 & 5 & \textbf{15.85} & 26.30 & \textbf{14.35} & 28.40 & \textbf{14.25} & 20.35 & 12.15 & \textbf{26.30} \\
\bottomrule
\end{tabular}
}
\end{table}

\rparagraph{Gap between Zero-shot and Few-shot Learning}
In contrast to previous results on text-only tasks, we find few-shot learning for multilingual multimodal tasks to be especially challenging, possibly due to the complex nature of the respective tasks combined with the sparsity of the few-shot learning data. 
As plotted in \Cref{fig:fewshot-curves-avgV2}, we observe no clear improvement as the number of shots increase in all tasks, the only exception being xGQA. 
This suggests that cost-efficiently hand-labelling $\sim100$ shots in order to improve the models performance on the target language, unlike in text-only setups, is insufficient to make meaningful progress on \vl tasks such as XVNLI, MaRVL and xFlickr\&Co. 
This also stresses the importance of developing methods for 
\emph{improved} zero-shot transfer in future work, as \vl models still require sufficient amounts of expensive task-annotated data to benefit from in `non-zero-shot' setups~(\cref{tab:few-shot-diffs}).

In contrast to XVNLI, MaRVL and xFlickr\&Co, we find that few-shot experiments on xGQA yield large performance gains in the target language. 
We attribute this to the structured nature of the data and the large label space of the task (1{,}842 classes) resulting in a misaligned multilingual embedding space in zero-shot setups, as suggested by the original authors. 
We analyse these results further in \S\ref{sec:analysis}.

\rparagraph{Increasing the Number of Shots} 
In the few-shot experiments, we noticed no observable improvement or clear trend of performance increase on 3 out of the 4 tasks. 
In order to identify if, and at what point, increasing the number of shots would start being beneficial, we extend our few-shot experiments on the retrieval task using the xFlickr\&CO dataset. 
In \cref{fig:fewshot-curves-xflickr-deja}, we plotted \german{} and \japanese{} performance with up to 1{,}500 shots\footnote{\german{} and \japanese{} captions are from Multi30K~\cite{elliott-etal-2016-multi30k} and \citet{nakayama-etal-2020-visually}, respectively.} (instead of the 100 shots used in our standard few-shot setup for this dataset) for the two best performing models in this task.
Indeed, we observe an overall trend of increasing performance with more training examples, although with noticeable instability issues (e.g. UC$^2$ accuracy in \japanese{} drops when more than 500 shots are used).
These results demonstrate that more training examples may indeed further improve the overall performance in the target language, but the large number of examples required questions whether this setup can even be dubbed `few-shot'.

\rparagraph{More Images versus More Captions} 
An important question, which concerns data collection efficiency for the target languages in image--text retrieval scenarios, is whether it is more useful, for a given budget, to collect (1) more \emph{captions per image} or (2) more \emph{images with a single caption}. 
To investigate this question, we trained xUNITER and UC$^2$ on \japanese{} and \german{} data with different number of shots and different ratios of captions per image
(\Cref{tab:FlickrCO-num-ir}). 
As indicated by the results, there is no clear trend suggesting that one setting is better than the other. We leave a larger-scale investigation related to this question for future work. 

\section{Further Analyses} \label{sec:analysis}

We now provide a series of further and finer-grained analyses, which include performance of the `translate train' transfer approach and domain-specific results in xFlickr\&CO, as well as an analysis of performance over structurally different question types of xGQA in few-shot setups.


\rparagraph{Translate Train Performance}
We further compare the few-shot results with `translate train' performance: We directly fine-tune the pretrained UC$^2$ and xUNITER checkpoints on the full, human-translated Flickr30K training set. 
As shown in \cref{fig:fewshot-curves-xflickr-deja}, UC$^2$ achieves remarkable zero-shot and few-shot performance: 100 or fewer data points are enough for it to match `translate train' performance, which it surpasses with 1{,}000 or more data points.
Similar results can also be observed in Japanese, although it needs 500 data points to match `translate train' accuracy.
xUNITER---which was not pretrained on German nor Japanese multimodal data---also reaps benefits from 500 or more data points, although still being far from its `translate train' counterpart (despite being initialised from English fine-tuned weights).
Finally, we note that `translate train' performance is still far from the corresponding English performance.
This result is rather surprising for UC$^2$, which was solely pretrained on Conceptual Captions in five languages, each of equal size.

\begin{figure}[t]
 	\centering
 	\includegraphics[width=\linewidth, trim={0cm 0cm 0cm 0cm}, clip]{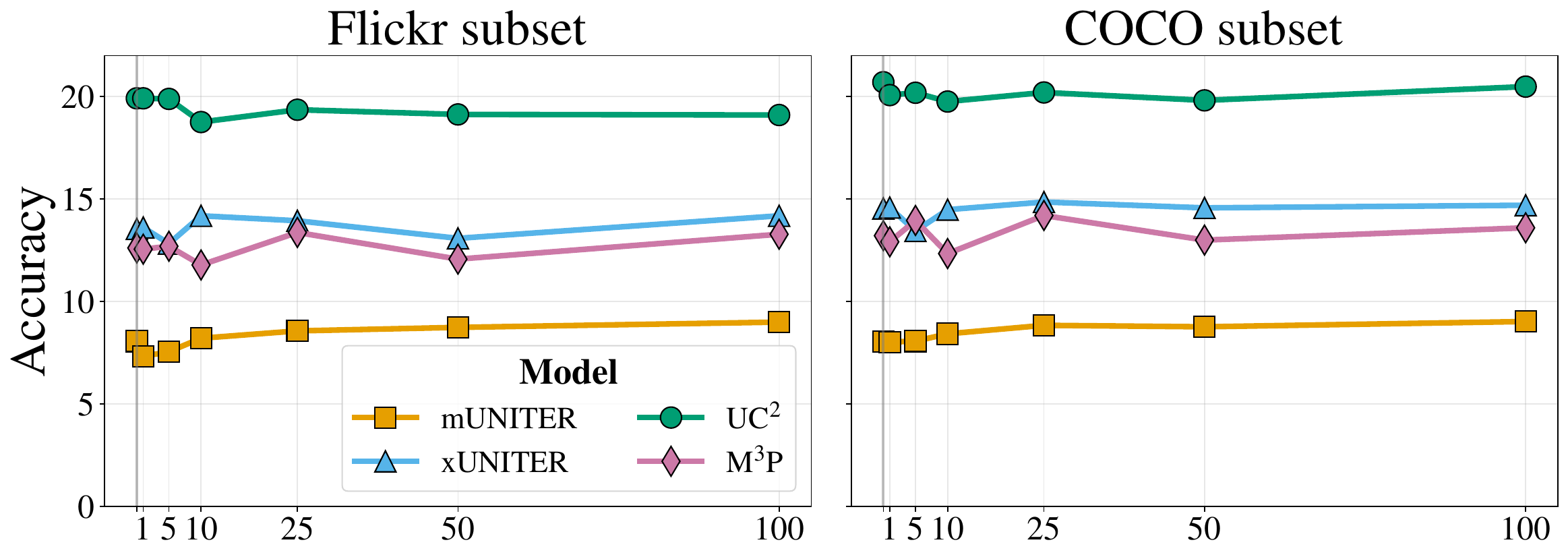}%
 	\vspace{-0.2cm}
 	\caption{Image retrieval performance averaged across languages per xFlickr\&CO subset.}
 	\label{fig:fewshot-curves-xflickr-subset}
\vskip 0.15in
 	\centering
 	\includegraphics[width=\linewidth, trim={0cm 0cm 0cm 0cm}, clip]{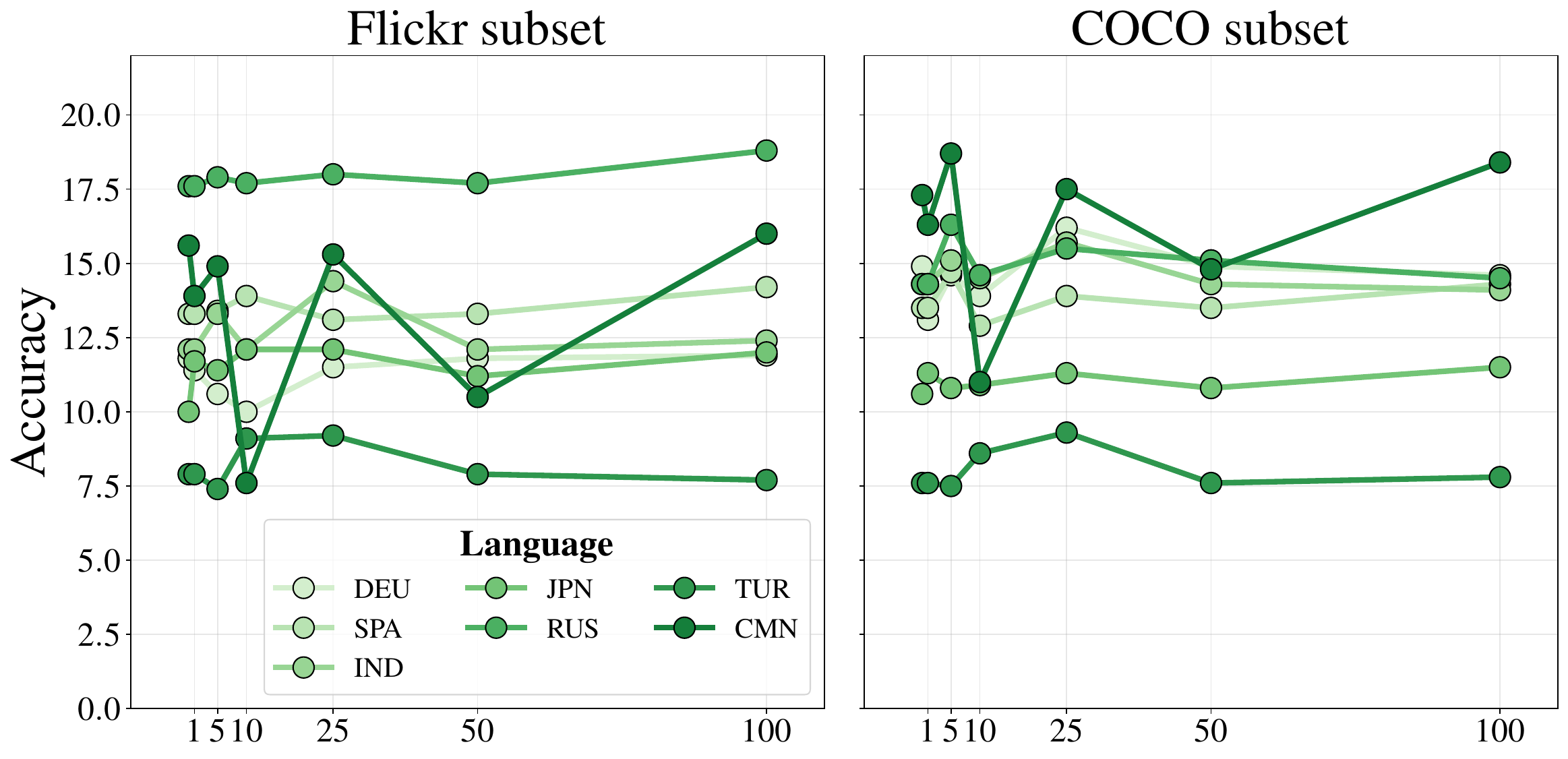}%
 	\vspace{-0.2cm}
 	\caption{UC$^2$ image retrieval performance across languages per xFlickr\&CO subset.}
 	\label{fig:fewshot-curves-uc2-xflickrco-subset}
\end{figure}

\begin{table}[t]
\caption{English image retrieval performance by evaluation subset in xFlickr\&CO after fine-tuning.} 
\label{tab:xflickrco-subset}
\setlength\tabcolsep{3pt}
\small
\center
  \begin{tabular}{lrrrr}
\toprule
\textbf{xFlickr\&CO subset} & \multicolumn{1}{c}{\textbf{mUNITER}} & \multicolumn{1}{c}{\textbf{xUNITER}} & \multicolumn{1}{c}{\textbf{UC$^2$}} & \multicolumn{1}{c}{\textbf{M$^3$P}} \\ 
\midrule
Flickr & \textbf{51.3} & \textbf{46.6} & \textbf{44.0} & \textbf{37.1} \\
COCO & 37.7 & 30.3 & 30.8 & 25.6 \\
\bottomrule
\end{tabular}
\end{table}

\rparagraph{Retrieval Performance by Evaluation Subset}
Our xFlickr\&CO dataset is a composition of Flickr and COCO data (each of 1{,}000 image--sentence pairs).
As models are fine-tuned on Flickr, the COCO evaluation subset might require out-of-distribution generalisation due to its different visual domain.
Indeed, this is what we observe in English (\cref{tab:xflickrco-subset}).
However,~\cref{fig:fewshot-curves-uc2-xflickrco-subset} shows that this is not the case for the multilingual encoders, which perform similarly in both subsets.
A closer look at UC$^2$ further shows that this behaviour is language-independent.
Given the relatively poor performance of current models, we cannot assess whether the domain shift we observe in English does not happen in xFlickr\&CO due to its unified captioning guidelines, or whether domain generalisation will be a challenge in xFlickr\&CO for future, tightly-multilingual models.